\documentclass[10pt,journal,compsoc]{IEEEtran}

\usepackage{times}
\usepackage{epsfig}
\usepackage{graphicx}
\usepackage{amsmath}
\usepackage{amssymb}
\usepackage{dsfont}
\usepackage{siunitx}
\usepackage{subfigure}
\usepackage{multirow}
\usepackage{caption}
\usepackage{xcolor}
\usepackage{comment}

\usepackage[pagebackref=true,breaklinks=true,colorlinks,bookmarks=false]{hyperref}

\def\x{\textbf{x}}
\def\X{\textbf{X}}
\def\y{\textbf{y}}

\def\C{\textbf{C}}
\newcommand{\myparagraph}[1]{\vspace{2pt}\noindent{\bf #1}}

\begin{document}

\title{Generalized Few-Shot Video Classification with Video Retrieval and Feature Generation}

\author{Yongqin~Xian, Bruno~Korbar, Matthijs~Douze, Lorenzo~Torresani, \\ Bernt~Schiele and Zeynep~Akata
\IEEEcompsocitemizethanks{
\IEEEcompsocthanksitem Yongqin Xian is with ETH Zurich. The majority of the work was done when Yongqin Xian was with Max Planck Institute for Informatics, Germany.
\IEEEcompsocthanksitem Bruno~Korbar is with University of Oxford.
\IEEEcompsocthanksitem Matthijs~Douze is with Facebook.
\IEEEcompsocthanksitem Lorenzo~Torresani is with Facebook and with Dartmouth College, USA.
\IEEEcompsocthanksitem Bernt Schiele is with Max Planck Institute for Informatics, Germany.
\IEEEcompsocthanksitem Zeynep~Akata is with University of T\"ubingen, Max Planck
Institute for Informatics and Max Planck Institute for Intelligent Systems, Germany. 
}
\thanks{}}
\IEEEtitleabstractindextext{%
\begin{abstract}
%1. use 3D CNN, simple baseline, outperform
%2. two extensions 
%3. more challenging benchmarks
Few-shot learning aims to recognize novel classes from a few examples. Although significant progress has been made in the image domain, few-shot video classification is relatively unexplored. We argue that previous methods underestimate the importance of video feature learning and propose to learn spatiotemporal features using a 3D CNN. Proposing a two-stage approach that learns video features on base classes followed by fine-tuning the classifiers on novel classes, we show that this simple baseline approach outperforms prior few-shot video classification methods by over 20 points on existing benchmarks. To circumvent the need of labeled examples, we present two novel approaches that yield further improvement. First, we leverage tag-labeled videos from a large dataset using tag retrieval followed by selecting the best clips with visual similarities. Second, we learn generative adversarial networks that generate video features of novel classes from their semantic embeddings. Moreover, we find existing benchmarks are limited because they only focus on 5 novel classes in each testing episode and introduce more realistic benchmarks by involving more novel classes, i.e. few-shot learning, as well as a mixture of novel and base classes, i.e. generalized few-shot learning. The experimental results show that our retrieval and feature generation approach significantly outperform the baseline approach on the new benchmarks. Code will be  available at \url{https://github.com/xianyongqin/few-shot-video-classification}. 
\end{abstract}

}

\maketitle
\IEEEdisplaynontitleabstractindextext
\IEEEpeerreviewmaketitle

\section{Introduction}

%Limitations of existing approaches: 1. strong video feature representation is not considered or inappropriately used

%Discuss our high-level ideas of two-stage baseline 

%Discuss video retrieval (explore the abundant tag-labeled videos available in the real world)

%Discuss feature generation (multi-modal transfer learning). 

%benchmark not sufficiently challenging

%Summarize contributions

\begin{figure}[t]
	\centering
	\includegraphics[width=\linewidth, trim=0 0 0 0,clip]{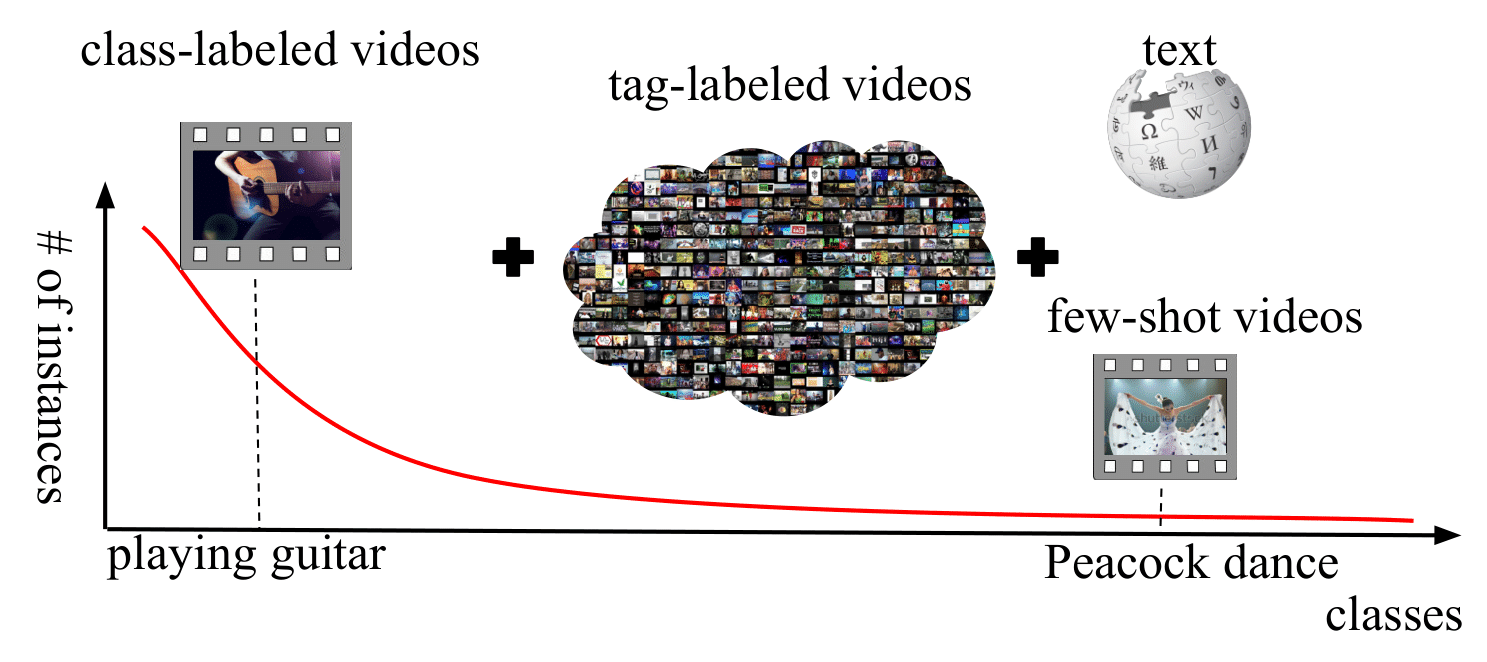}
	\vspace{-5mm}
	\caption{Our 3D CNN approach combines a few class-labeled videos (time-consuming to obtain) and tag-labeled videos. %It saturates existing benchmarks, so we move to a more challenging generalized many-way few-shot video classification task. %Faces in the figure are artificially masked for privacy reasons. 
	}
	\vspace{-5mm}
	\label{fig:teaser}
\end{figure}

%Compared to humans, machine learning algorithms require far more annotated training data to achieve a good performance. 
Humans have the remarkable ability to learn a novel concept from only a few observations. In contrast, machine learning algorithms still require a large amount of labeled data to achieve a good performance. 
This is problematic because collecting a big amount of labeled data is not always an easy task. In the video domain, annotating data is particularly challenging due to the additional time dimension. Hence, the lack of labeled training data is more prominent for some fine-grained action classes at the ``tail'' of the skewed long-tail distribution (see Figure~\ref{fig:teaser}, ``peacock dance''). %\zeynep{I thought your activities are only for humans. This example gives the impression that you can generalize to any class. I would replace this example with an actual action example from the dataset.} \yongqin{peacock dance is a dancing style of human. it is an example of rare action classes. unfortunately there is no rare action class in the standard dataset.} 
It is thus important to study learning algorithms that generalize well to novel classes with only limited number of labeled training data, which is known as few-shot learning. While considerable attention has been devoted to this scenario in the image domain~\cite{vinyals2016matching,qi2018low,ravi2016optimization,chen2019closer}, few-shot video classification has received far less attention.

%More specifically, visual recognition methods that operate in such few-shot learning setting aim to generalize a classifier trained on base classes i.e., with enough training data to novel classes with only a few labeled training examples.

The goal of few-shot video classification is to predict the action labels of videos given only a few training examples of those classes. Compared with image classification, this has another dimension of complexity because we have to properly model the temporal information in the videos. There have been several attempts~\cite{zhu2018compound,cao2019few,bishay2019tarn} to tackle this challenging problem. These methods follow the meta-learning principle~\cite{vinyals2016matching, ravi2016optimization} to ``meta-learn'' a compound memory network~\cite{zhu2018compound} or a distance metric for comparing video similarities~\cite{cao2019few, bishay2019tarn}. However, we argue that those approaches~\cite{zhu2018compound,cao2019few,bishay2019tarn} underestimate the importance of video feature representation and ignore the imbalanced issue between base classes i.e., classes with many
training instances, and novel classes i.e., classes with only a few training instances. In particular, CMN~\cite{zhu2018compound} and TAM~\cite{cao2019few} are built on frame-level features extracted from a 2D CNN, which essentially ignores the important temporal information. 
%Although additional temporal modules have been added at the top of a pre-trained 2D CNN, necessary temporal cues may be lost when temporal information is learned on top of static image features.
We argue that under-representing temporal cues may negatively impact the robustness of the classifier. In fact, in the few-shot scenario it may be risky for the model to rely exclusively on appearance and context cues extrapolated from the few-shot available examples.
Although TARN~\cite{bishay2019tarn} adopts the stronger C3D~\cite{tran2015learning} that is pre-trained to extract the  video features, it only achieves marginal performance boost over CMN~\cite{zhu2018compound} because it ignores the substantial domain gap between the pre-training dataset and the target dataset. 

%In this work, we propose to improve the video feature representation for few-shot video classification by utilizing the state-of-the-art 3D CNN i.e., R(2+1)D~\cite{tran2018closer} as the backbone. The R(2+1)D directly takes input as a short video clip and learn the temporal information that is important for distinguishing the target action classes. We then revisit a simple baseline with a two-stage training scheme: the first stage performs the representation learning on base classes and the second stage fine-tunes the classifiers for novel classes. This simple baseline surprisingly outperforms the state-of-the-art methods by a wide margin, indicating the importance of video feature representation for few-shot video classification.

%Furthermore, we introduce two approaches to extend the two-stage baseline.  alleviate the lack of labeled training data.   
While labeled videos of novel classes are sparse, there are abundant videos tagged by users available on the internet. For example, there are 800,000 videos in the YFCC100M~\cite{thomee2015yfcc100m} dataset. A subset of those videos might be related to our target novel classes, providing more training videos that cover diverse variations of the classes. However, mining this subset of relevant videos in a large-scale dataset is hard because videos in the dataset can have arbitrary lengths and most of the videos are irrelevant. Although videos are annotated with user tags, they are often not reliable due to the substantial tag noise. Our second goal is thus to leverage those tag-labeled videos~(Figure~\ref{fig:teaser}) by proposing efficient retrieval methods to find in the large-scale dataset e.g., YFCC100M, video clips related to the novel classes. The retrieved video clips serve as auxiliary training data to improve classifiers for the novel classes.  
%\zeynep{This paragraph is too application specific. I would not mention dataset names as those datasets may not be obvious to the reader and you haven't provided any dataset statistics. Keep your narrative general.}

%to alleviate the lack of labeled training data for novel classes. 

%This is different from webly supervised learning~\cite{chen2015webly, divvala2014learning} where they often rely on the output of powerful search engines like Google.

%Several methods~\cite{chen2015webly, divvala2014learning} have attempted to make use of those millions of images or videos online by obtaining training images from search engines like Google. However, the machine learning models behind the search engines might have been pre-trained on the novel classes, which violates the few-shot learning assumption.
However, our target classes e.g., action class ``pulling something from right to left'' are not always covered by the large-scale dataset e.g., YFCC100M. 
%the YFCC100M dataset does not always cover videos that are relevant to our target classes e.g., action class ``pulling something from right to left'' \zeynep{same as above. keep your narrative general, without mentioning dataset names in the main argument. You can give the dataset in an example, after e.g.}.
In this case, augmenting the dataset with synthesized variations of the training data conditioned on novel classes becomes a good alternative. Although significant progress~\cite{gulrajani2017improved, karras2019style} has been achieved in image generation, generating videos~\cite{tulyakov2018mocogan, vondrick2016generating} that can be used for training a deep learning architecture remains to be a challenging problem because the high dimensionality of the output space. Nevertheless, we realize that generating low-dimensional video features might be sufficient since our goal is to achieve good classification performance rather than generating realistic raw videos. Similar to image generation, video feature generation can be achieved by learning a generative model that allows us to sample new video features from the learned distribution. To generate video features that are complementary to the few-shot reference examples, we propose to rely on the semantic embeddings of class names i.e.,  Word2Vec~\cite{mikolov2013distributed} or BERT~\cite{devlin2018bert}, to control the modes of the generation. Our key insight is that the semantic embeddings can encourage knowledge transfer between base and novel classes because the semantic embedding space captures class similarities from the language modality. %

%To this end, we propose to extend the generative adversarial networks~(GANs)~\cite{WG16, gulrajani2017improved} to generate video features of novel classes in the CNN feature space i.e., R(2+1)D~\cite{tran2018closer}. In order to generate video features that are complementary to the few-shot examples, we control the modes of generated features using the semantic embeddings of class names i.e.,  Word2Vec~\cite{mikolov2013distributed} or BERT~\cite{devlin2018bert}, as the condition variables of the GANs. This 
%Although similar ideas have been explored in the fields of zero-shot image classification~\cite{xian2018feature}, for the video domain.

Furthermore, we find that existing experimental settings~\cite{zhu2018compound,cao2019few} for few-shot video classification have several limitations. Specifically, current benchmarks: 1) consider only 5 novel classes in each testing episode i.e., the model only predicts 5 novel classes. In contrast, there are far more novel classes to distinguish in the real world. 
%Predicting a label among just 5 novel classes is in fact relatively easy. 
2) ignore base classes at the test time i.e., the label space is restricted to include only novel classes. However, this is unrealistic because test videos are expected to belong to any class in real-world applications. We argue that a more challenging and realistic benchmark is required to push the progress of the field.

The main contributions of this work are: 1) We propose to improve the video features for few-shot video classification by adopting the state-of-the-art 3D CNN i.e., R(2+1)D~\cite{tran2018closer}. We then develop a simple baseline~(namely TSL) with a two-stage learning scheme: the first stage performs the representation learning on base classes and the second stage fine-tunes the classifiers for novel classes. This simple baseline surprisingly outperforms the state-of-the-art methods by a wide margin, indicating the importance of video feature representation for few-shot video classification.  2) We propose to leverage tag-labeled videos from a large dataset using tag retrieval followed by selecting the best clips with visual similarities, yielding further improvement; 3) As a complementary alternative, we develop a feature generating networks~(namely VFGAN) that generate video features from semantic embeddings, expanding the training set of novel classes to build better classifiers. 4) We extend current evaluation settings by introducing two challenges. In generalized few-shot video classification setting, the evaluation protocol expands the label space to include all the classes and reports base class accuracy, novel class accuracy and their harmonic mean. In many-way few-shot video classification setting, the number of novel classes goes beyond five, and towards all available classes. 5) Our approaches establish a new state-of-the-art on the existing benchmarks, improving prior best methods by over $20\%$ on two datasets i.e., Kinetics and SomethingV2. On the proposed challenging benchmarks,  our retrieval method, feature generation approach and their combination significantly outperform the baseline.

\section{Related work}
%\zeynep{I am skipping the related work, will come back if I have time}
\myparagraph{Few-shot learning setup.}
The few-shot image classification~\cite{ravi2016optimization,hariharan2017low} setting uses a large-scale fully labeled dataset for pre-training a DNN on the base classes, and a few-shot dataset with a small number of examples from a disjoint set of novel classes. 
The terminology ``$k$-shot $n$-way classification'' means that in the few-shot dataset there are $n$ distinct classes and $k$ examples per class for training.
Evaluating with few examples ($k$ small) is bound to be noisy. 
Therefore, the $k$ training examples are often sampled several times and accuracy results are averaged~\cite{hariharan2017low,douze2018low}. 
Many authors focus on cases where the number of classes $n$ is small as well, which amplifies the measurement noise. 
For that case~\cite{ravi2016optimization} introduces the notion of ``episodes''. An episode is one sampling of $n$ classes and $k$ examples per class, and the accuracy measure is averaged over episodes. 

It is feasible to use distinct datasets for pre-training and few-shot evaluation. 
However, to avoid dataset bias~\cite{torralba2011unbiased} it is easier to split a dataset into disjoint sets of ``base'' and  ``novel'' classes.
The evaluation is  often performed only on novel classes, except \cite{hariharan2017low} who evaluate on the combination of base and novel classes. Recently, few-shot video classification benchmarks have been proposed~\cite{zhu2018compound,cao2019few}. 
They use the same type of decomposition of the dataset % \zeynep{how can they use the same split in a video dataset as a few-shot image classification method trained on images?? the datasets can not be the same.} 
as~\cite{ravi2016optimization}, with learning episodes and random sampling of novel classes. 
In this work, we extend the benchmarks of~\cite{zhu2018compound, cao2019few} by including more challenging settings.

\myparagraph{Tackling few-shot learning.}
The simplest few-shot learning approach is to extract embeddings from the images using the pre-trained DNN and train a linear classifier~\cite{APHS15} or logistic regression~\cite{hariharan2017low} on these embeddings using the $k$ available training examples. Other notable works include adopting a nearest-neighbor classifier~\cite{Wang2019SimpleShot} and learning to ``imprinting'' a linear classifier from training examples~\cite{qi2018low}.
As a complementary approach, \cite{joulin2016learning} has looked into exploiting tags of 100M images from the YFCC100M dataset~\cite{thomee2015yfcc100m} to aid classification.
%By leveraging tags of 100M images from the YFCC100M dataset~\cite{thomee2015yfcc100m}
%, they show improvements over Imagenet-pretraining. %ed classifiers on transfer learning tasks.
In this work, we use videos from YFCC100M retrieved by tags to augment and improve training of our few-shot classifiers. 

%Another approach is to cast low-shot learning as a nearest-neighbor classifier~\cite{Wang2019SimpleShot}. The ``imprinting'' approach~\cite{qi2018low}, consists in building a linear classifier from the embeddings of training examples, then fine-tune it. Note that this is close to a nearest-neighbor classifier, since it is equivalent to doing class-mean similarity search with a cosine distance. 

%ICI~\cite{wang2020instance}: Its main idea is to iteratively train a classifier, compute pseudo-labels for unlabeled data, and select trustworthy pseudo-labeled instances alongside the labeled examples to re-train the classifier. 

%Learning to propagate labels: Transductive propagation network for few-shot learning~\cite{liu2018learning}

%Learning to Self-Train for Semi-Supervised Few-Shot Classification~\cite{NEURIPS2019_bf25356f}

%Low-shot learning from imaginary data~\cite{wang2018low}

%Decoupling representation and classifier for long-tailed recognition~\cite{kang2019decoupling}

%Few-shot learning via embedding adaptation with set-to-set functions~\cite{ye2020few}: that takes all instances from the few-shot support set and outputs the set of adapted support instance embeddings, with elements in the set co-adapting with each other.

%\myparagraph{Meta-learning.} 
In a meta-learning setup, the few-shot classifier is assumed to have hyper-parameters or parameters that must be adjusted before training. 
Thus, there is a preliminary meta-learning step that consists in training those parameters on simulated episodes sampled from the base classes. 
Both Matching networks~\cite{vinyals2016matching} and Prototypical Networks\cite{snell2017prototypical} employ metric learning to ``meta-learn''  deep neural features and adopt a nearest neighbor classifier. 
%\cite{ravi2016optimization} meta-learns  the optimization algorithm via an LSTM that maps the low-shot training examples into a classifier. 
Feature hallucination~\cite{hariharan2017low,wang2018low} meta-learns how to generate additional training data for novel classes, directly in the feature space. Recently, \cite{ye2020few} propose to learn task-specific instance embeddings with the Transformer architecture~\cite{vaswani2017attention}.  
%In MAML~\cite{finn2017model}, the embedding classifier is meta-learned to adapt quickly and without overfitting to fine-tuning. 
%Ren et al.~\cite{ren2018meta} introduce a semi-supervised meta-learning approach that include unlabeled examples in each training episode. 
Semi-supervised meta-learning approaches~\cite{ren2018meta,liu2018learning, NEURIPS2019_bf25356f, wang2020instance} have shown that including unlabeled examples in each training episode is beneficial for few-shot learning. Their key idea is to iteratively train a classifier, compute pseudo-labels for unlabeled data, and select trustworthy pseudo-labeled instances alongside the labeled examples to re-train the classifier. While these methods hold out a subset from the same target dataset as the unlabeled images, in our setup, the tag-labeled videos from a large-scale heterogeneous dataset may have domain shift issues and a huge amount of distracting videos. Moreover, prior pseudo-labeling algorithms are computationally expensive, hindering them from leveraging a large dataset.

Recent works~\cite{chen2019closer,Wang2019SimpleShot,kang2019decoupling} suggest that state-of-the-art performance can be obtained by methods that do not need meta learning.
In particular, \cite{chen2019closer} show that meta-learning methods are less useful when the image descriptors are expressive enough, which is the case when they are from high-capacity networks trained on large datasets. Therefore, we focus on techniques that do not require a meta-learning stage. 

\myparagraph{Deep descriptors for videos.}
Moving from hand-designed descriptors~\cite{wang2013action} to learned deep network based descriptors~\cite{feichtenhofer2016spatiotemporal, karpathy2014large,simonyan2014two,wang2016temporal,tran2015learning} has been enabled by labeled large-scale datasets~\cite{kay2017kinetics,karpathy2014large}, and parallel computing hardware. Deep descriptors are sometimes based on 2D-CNN models operating on a frame-by-frame basis with temporal aggregation~\cite{girdhar2017actionvlad}.
More commonly they are 3D-CNN models that operate on short sequences of images that we refer to as video-clips~\cite{tran2015learning,tran2018closer}.
Recently, ever-more-powerful descriptors have been developed leveraging two-stream architectures using additional modalities~\cite{feichtenhofer2016convolutional,simonyan2014two}, factorized 3D convolutions~\cite{tran2018closer}, or multi-scale approaches~\cite{feichtenhofer2019slowfast}. %While most of these descriptors are trained in a fully supervised way, advances in learning deep descriptors in either weakly supervised~\cite{yalniz2019billion,ghadiyaram2019large} or self supervised fashion have been explored as well~\cite{korbar2018cooperative,owens2018audio}.

\myparagraph{Generative Adversarial Networks.}
Generative Adversarial Networks~(GANs)~\cite{GPMXWDOCB14} were originally developed as a generative model to learn arbitrary data distributions e.g., images from a particular domain. 
%The architecture of GANs consists of a generator and a discriminator that compete with each other in a two-player minimax game. 
In the context of image generation, the generator takes input as a random noise vector and produces a generated image, while the discriminator tries to distinguish real images from generated ones. 
%GANs have attracted increasing attention and significant progress has been achieved recently~\cite{WG16, gulrajani2017improved, karras2019style, choi2018stargan}. 
Many prior works focus on solving the instability issues of GANs. In particular, the theory of GANs has been studied in \cite{arjovsky2017towards} and WGAN~\cite{gulrajani2017improved, arjovsky2017wasserstein} that optimizes the Wasserstein distance has been proposed to cure the unstable issues. 
%Arjovsky et al.~\cite{arjovsky2017towards} show that the Jenson-Shannon divergence optimized by the original GAN leads to instability issues and propose Wasserstein GAN~(WGAN) that optimizes the approximation of Wasserstein distance~\cite{arjovsky2017wasserstein}.
%Gulrajani et al.~\cite{gulrajani2017improved} improve WGAN~\cite{arjovsky2017wasserstein} by using a gradient penality loss to enforce the smoothness constraints on the discriminator. 
Besides, adding additional loss terms seem to improve GANs training as well e.g., the mutual information loss~\cite{infogan} and the VAE loss~\cite{larsen2016autoencoding}. %Another group of works focus on improving the discriminator by using multiple discriminators~\cite{doan2019line} or self-attention~\cite{zhang2019self}. 
%There are also some works that study the input latent space of the generator by using mixture of Gaussians~\cite{ben2018gaussian}, hierarchical latent spaces~\cite{brock2018large}, or clustering~\cite{mukherjee2019clustergan}.
%Progressive growing GANs~\cite{karras2017progressive} synthesize high-resolution images by progressively expanding the image resolution. 
Recently, styleGAN~\cite{karras2019style} introduces a novel generator architecture that includes an adaptive instance normalization and a mapping network at each intermediate layer of the generator. 
 
GANs have shown a wide range of applications. Conditional GAN~\cite{conditionalgans} is able to control the generator to produce samples from a particular class by including conditional variables e.g., class labels, to the generator and discriminator. In addition, GANs may also be conditioned on more complex types of data, such as natural language sentences describing the content of the image~\cite{RAYLSL16}. Apart from synthesizing realistic looking images~\cite{ karras2019style}, GANs could also be used for unsupervised representation learning~\cite{RMC16} and video generation~\cite{vondrick2016generating,tulyakov2018mocogan}. Recently, GANs have been applied to generate CNN image features~\cite{xian2018feature, zhu2018generative} of unseen classes for zero-shot image classification. Specifically, f-CLSWGAN~\cite{xian2018feature} developed a GAN-based model that synthesizes CNN image features conditioned on the class-level attributes or word embeddings. They~\cite{xian2018feature, zhu2018generative} show that the generative models learned on seen classes generalize well to generate image features of unseen classes by feeding their class attributes to the learned generator.        
 
In this work, we want to apply GANs to augment the few-shot video classification task where the novel classes have only a limited number of training videos. We argue that generating high-quality videos is a hard task and therefore propose to generate video features which have a much smaller dimensionality. Although similar ideas have been explored in \cite{xian2018feature} and \cite{zhu2018generative} in the context of zero-shot learning image classification~\cite{xian2018zero}, our work is to study video feature generation for few-shot video classification, which is a harder task as the video features have to encode the critical temporal information.

\myparagraph{Video Retrieval.} In general, the goal of video retrieval is to search videos of the same actions specified by query videos. Many previous works~\cite{ Baraldi_2018_CVPR,douze2016circulant} represent a video as a collection of image features extracted from frames or keyframes and compare the query descriptors with  those of the videos from the database. In order to retrieve temporal-consistent videos, temporal constraints are enforced by the temporal match kernels~\cite{Baraldi_2018_CVPR}. But those approaches often do not scale well to the large-scale video database due to the heavy computational cost. In addition, in the few-shot learning scenario, another challenge for video retrieval is that the target action class has only a limited number of query examples. To address this issue, some works~\cite{mazloom2014few,agharwal2016tag} leverage video tags to provide complementary information by performing tag propagation followed by tag retrieval~\cite{mazloom2014few}, and mapping video descriptors to the word embedding space~\cite{agharwal2016tag}.

\section{Few-Shot Learning with Video Retrieval and Video Feature Generation}
%\section{Approaches \zeynep{too generic of a title. Make it specific. Which approaches are you talking about?}}

\begin{figure*}[t]
	\centering
		\includegraphics[width=0.9\textwidth, trim=0 0 0 0,clip]{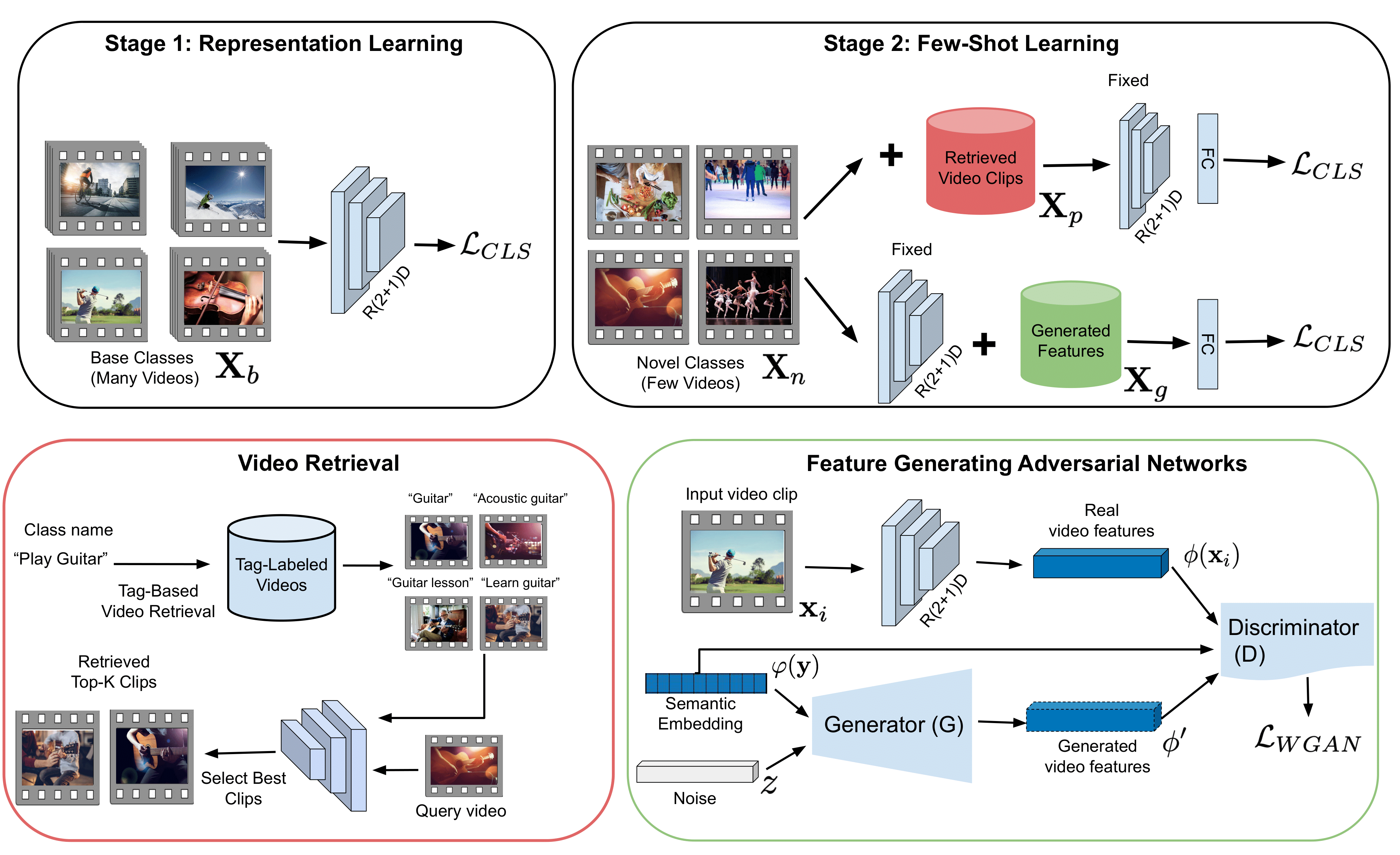}
	\caption{Our approach comprises two learning stages: representation learning and few-shot learning. In representation learning, we train a R(2+1)D CNN using the base classes starting from a pretrained model. In few-shot learning, %given few-shot support videos from novel classes, 
	we propose to expand the few-shot training set of novel classes by either retrieving videos from another tag-labeled video dataset or generating video features from semantic embeddings. Our video retrieval algorithm first estimates a list of candidate videos for each class from YFCC100M~\cite{thomee2015yfcc100m} using their tags, followed by selecting the best matching short clips from the retrieved videos using visual features. Those clips serve as additional training examples to learn classifiers that generalize to novel classes at test time. 
	Moreover, we propose to learn the feature generator that sythesizes video features from a semantic embedding with generative adversarial networks. 
	%and a random noise while the discriminator tries to distinguish generated video features from real features. %Both the  generator and discriminator are implemented with MLP and learned by optimizing the Wasserstein GAN loss. 
 }
	\label{fig:approach}
\end{figure*}

In the few-shot learning setting~\cite{zhu2018compound}, classes are split into two disjoint label sets, i.e., base classes~(denoted as $\C_b$) that have a large number of training examples, and novel classes~(denoted as $\C_n$) that have only a small set of training examples. Let $\X_{b}$ denote the training videos with labels from the base classes and $\X_{n}$ be the training videos with labels from the novel classes~($|\X_{b}| \gg |\X_{n}|$). 
Given the training data $\X_b$ and $\X_n$, the goal of the conventional few-shot video classification task ~(FSV)~\cite{zhu2018compound,cao2019few} is to learn a classifier which predicts labels among novel classes at test time. As the test-time label space is restricted to a few novel classes, the FSV setting is unrealistic. Thus, in this paper, we additionally study the generalized few-shot video classification~(GFSV) which allows videos at test time to belong to any base or novel class. 

\subsection{Two-Stage Learning Baseline}
\label{sec:3dfsv}
In this section, we introduce a simple two-stage learning baseline~(TSL) for few-shot video classification. Our approach in Figure~\ref{fig:approach} consists of 1) a representation learning stage which trains a spatiotemporal CNN i.e., R(2+1)D~\cite{tran2018closer}, on the base classes, 2) a few-shot learning stage that trains a linear classifier for novel classes with few labeled videos. %The details of each of these stages are given below.

\myparagraph{Representation learning.} Our model adopts a 3D CNN~\cite{tran2018closer} 
$\phi:\mathbb{R}^{F\times 3\times H\times W}\rightarrow \mathbb{R}^{d_v}$, encoding a short, fixed-length video clip of $F$ RGB frames with spatial resolution $H\times W$ to a feature vector in a $d_v$-dimensional embedding space. On top of the feature extractor $\phi$, we define a linear classifier $f(\bullet;W_b)$ parameterized by a weight matrix $W_b\in \mathbb{R}^{d_v\times |\C_b|}$, producing a probability distribution over the base classes. The objective is to jointly learn the network $\phi$ and the classifier $W_b$ by minimizing the cross-entropy classification loss on video clips randomly sampled from training videos $\X_b$ of base classes. More specifically, given a training video $\x\in \X_b$ with a label $\y\in \C_b$, 
the loss for a video clip $\x_i\in \mathbb{R}^{F\times 3\times H\times W}$ sampled from video $\x$ is defined as
\begin{equation}
\label{eq:rl}
     \mathcal{L}(\x_i) = -\log \sigma(W_b^T\phi(\x_i))_{\y}
\end{equation} 
where $\sigma$ denotes the softmax function that produces a probability distribution and $\sigma(\bullet)_{\y}$ is the probability at class $\y$.
 Following~\cite{chen2019closer}, we do not do meta-learning, so we can use all the base classes to learn the network $\phi$.

\myparagraph{Few-shot learning.} This stage aims to adapt the learned network $\phi$ to recognize novel classes $\C_n$ with a few training videos $\X_n$. To reduce overfitting, we fix the network $\phi$ and learn a linear classifier $f(\bullet, W_n)$ by minimizing the cross-entropy loss on video clips randomly sampled from videos in $\X_n$, where $W_n\in \mathbb{R}^{d_v\times |\C_n|}$ is the weight matrix of the linear classifier. Similarly, we define the loss for a video clip $\x_i$  sampled from $\x\in \X_n$ with a label $\y$ as
\begin{equation}
\label{eq:ce}
     \mathcal{L}(\x_i) = -\log \sigma(W_n^T\phi(\x_i))_{\y}
\end{equation} 

\myparagraph{Testing.} The spatiotemporal CNN operates on fixed-length video clips of $F$ RGB frames and the classifiers make clip-level predictions. At test time, the model must predict the label of a test video $\x \in \mathbb{R}^{T\times 3\times H\times W}$ with arbitrary time length $T$. We achieve this by randomly drawing a set $L$ of clips $\{\x_i\}_{i=1}^L$ from video $\x$, where $\x_i\in \mathbb{R}^{F\times 3\times H\times W}$. %denotes a video clip of successive $F$ frames and $L$ is the number of clips we process for each video. 
The video-level prediction is then obtained by averaging the prediction scores after the softmax function over those $L$ clips. For few-shot video classification (FSV), this is: 
% we predict the test video $\x$ as the winner-takes-all selected from the novel classes via computation of 
\begin{equation}
    \frac{1}{L}\sum_{i=1}^{L} f(\x_i; W_n).
\end{equation} 
For generalized few-shot video classification (GFSV), both base and novel classes are taken into account and we concatenate the base class weight $W_b$ learned in the representation stage with the novel class weight $W_n$ learned in the few-shot learning stage:
\begin{equation}
\frac{1}{L}\sum_{i=1}^{L} f(\x_i; [W_b; W_n]).
\end{equation} 

\subsection{Video Retrieval}% for few-shot action recognition}
During few-shot learning, fine-tuning the network $\phi$ or learning the classifier $f(\bullet;W_n)$ alone is prone to overfitting. Moreover, class-labeled videos to be used for fine-tuning are scarce. Instead, the hypothesis is that leveraging a massive collection of weakly-labeled real-world videos would improve our novel-class classifier. To this end, for each novel class, we propose to retrieve a subset of weakly-labeled videos, associate pseudo-labels to these retrieved videos and use them to expand the training set of novel classes. It is worth noting that those retrieved videos may be assigned with wrong labels and have domain shift issues as they belong to another heterogeneous dataset, making this idea challenging to implement. 
For efficiency and to reduce the label noise, we adopt the following two-step retrieval step.   

\myparagraph{Tag-based video retrieval.} The YFCC100M dataset~\cite{thomee2015yfcc100m} includes around 800K videos collected from Flickr, with a total length of over 8000 hours. %We treat the video hashtags, i.e. ``tags'', as a set of words. %
Processing a large collection of videos has a high computational demand and a large portion of them are irrelevant to our target classes. Thus, we restrict ourselves to videos with tags related to those of the target class names and leverage information that is complementary to the actual video content to increase the visual diversity. 

Given a video with user tags $\{t_i\}_{i=1}^S$ where $t_i\in \mathcal{T}$ is a word or phrase and $S$ is the number of tags, we represent it with an average tag embedding  $\frac{1}{S}\sum_{i=1}^{S} \varphi(t_i)$. We do not observe better performance when using max-pooling. The tag embedding $\varphi(.): \mathcal{T}\rightarrow \mathbb{R}^{d_t}$ maps each tag to a $d_t$ dimensional embedding space, e.g., Fasttext~\cite{joulin2016fasttext}. Similarly, we can represent each class by the text embedding of its class name and then for each novel class $c$, we compute its cosine similarity to all the video tags and retrieve the $N$ most similar videos according to this distance.

\myparagraph{Selecting best clips.} The video tag retrieval selects a list of $N$ candidate videos for each novel class. However, those videos are not yet suitable for training because the annotation may be erroneous, which can harm the performance. Besides, some weakly-labeled videos can last as long as an hour. We thus propose to select the best short clips of $F$ frames from those candidate videos using the few-shot videos of novel classes. 

Given a set of few-shot videos $\X_n^c$ from novel class $c$, we randomly sample $L$ video clips from each video. We then extract features from those clips with the spatiotemporal CNN $\phi$ and compute the class prototype by averaging over clip features. Similarly, for each retrieved candidate video of novel class $c$, we also randomly draw $L$ video clips and extract clip features from $\phi$. Finally, we perform a nearest neighbour search with cosine distance to find the $M$ best matching clips of the class prototype: %. This can be formulated as  
\begin{equation}
\max_{\x_j} \ \cos(p_c, \phi(\x_j))
\end{equation} 
where $p_c$ denotes the class prototype of class $c$, $\x_j$ is the clip belonging to the retrieved tag-labeled videos. After repeating this process for each novel class, we obtain a collection of pseudo-labeled video clips $\X_p=\{\X_p^c\}_{c=1}^{|C_n|}$ where $\X_p^c$ indicates the best $M$ retrieved video clips for novel class $c$. 

\myparagraph{Batch denoising.} The retrieved video clips contribute to learning a better novel-class classifier $f(\bullet; W_n)$ in the few-shot learning stage by expanding the training set of novel classes from $\X_n$ to $\X_n \bigcup \X_p$. However, it is  likely that $\X_p$ include video clips with wrong pseudo labels. During the optimization, we adopt a simple strategy to alleviate the noise: half of the video clips per batch come from $\X_n$ and another half from $\X_p$ at each iteration. The purpose is to reduce the gradient noise in each mini-batch by enforcing that half of the samples are trustworthy.

\subsection{Video Feature Generating Adversarial Networks}

The video retrieval step described in previous section requires a large collection of tag-labeled videos, which are not always available. In this section, we consider a data generation approach as an alternative to circumvent the data scarcity issue. 
%An important question is that what data we should generate. 
While one could apply some physics engines to render synthetic videos, but we have to address the domain gap issue which might be even more difficult.
Given the success of GANs on image and video generation~\cite{gulrajani2017improved, karras2019style,tulyakov2018mocogan, vondrick2016generating}, one might think about directly synthesizing training videos with GANs. However, generating high-quality videos with GANs remains a challenging problem because the dimensionality of the output space is high e.g., $16\times 3 \times 112\times 112$ if the video has 16 frames with spatial resolution $ 112\times 112$. 
To this end, we develop a novel video feature generating adversarial network~(VFGAN) that generates video features i.e., the top pooling units of a 3D CNN. Compared to video generation, video feature generation is an easier task to learn because the dimensionality of the feature space is much lower. 
%Although there have been several works that generate image features in the field of zero-shot image classification~\cite{xian2018feature}, extending it to generate video features for few-shot video classification is not trivial because the two tasks are fundamentally different. While image features only encode the visual information within a single image, video features need to additionally capture the temporal information across multiple image frames. Besides, zero-shot learning~\cite{xian2018feature} focuses on predicting the unseen classes i.e., classes that are not observed, based on their side information. In contrast, few-shot learning~\cite{zhu2018compound} concerns how to efficiently learn from a few training examples.   
%we are the first to generate features for few-shot video classification.  \zeynep{technically, any summary of the data is a feature, e.g. prototypes is a feature. So I would specify what I mean by 'feature' in the context of videos here (in the beginning of the section). Also, I imagine extending image feature generation to video feature generation is not that trivial so providing some specifics would be useful.} 
In the following, we first describe the semantic embedding followed by explaining how the feature generation is used in the few-shot learning stage. Finally, we discuss how the feature generator is learned. 

%Our approach is based on GANs which consist of a generator $G$ and a discriminator $D$ that compete with each other in a two-player game. In the context of image or video generation, $D$ aims to precisely distinguish real and generated images or videos, while $G$ tries to generate images or videos that fool $D$. GANs can be extended to be conditional GANs by feeding the conditions to both $G$ and $D$. In this work, rather than generating raw images or videos, we develop a variant of conditional GANs to generate clip-level features for novel classes. Rather than using class label as 

%our conditional GANs are able to generate highly discriminative clip-level features . To our knowledge, we are the first to apply GANs to synthesizes clip-level features for few-shot video classification.  

%\myparagraph{A discussion of why it is interesting and why it can work}

\myparagraph{Semantic embedding.} Our approach assumes the availability of some semantic embedding $\varphi(\y)\in \mathbb{R}^{d_y}$ that maps a class label to the $d_y$-dimensional embedding space in which class similarities can be captured e.g, ``dancing ballet'' is closer to ``dancing macarena'' than to ``playing drums''. In most of cases, the semantic embedding $\varphi(\y)$  can be obtained without extra annotations by extracting a word or sentence representation of the corresponding class name e.g., ``ballet'',  from a pretrained language model like Word2Vec~\cite{mikolov2013distributed} or BERT~\cite{devlin2018bert}. The semantic embedding has been widely used in zero-shot learning by learning a compatibility function between image and semantic embedding spaces~\cite{norouzi2013zero, elhoseiny2013write, frome2013devise}. In our work, the semantic embedding serves as the condition variables to our generative model so that it generates data for the desired classes. Besides, the language modality often includes information complementary to the visual modality and facilitate the knowledge transfer from base to novel classes.       

\myparagraph{Video feature generation for few-shot learning.} In the few-shot learning stage, we need to learn a classifier that predicts novel classes given a few labeled videos $\X_n$ from the novel classes and abundant labeled videos $\X_b$ from the base classes. However, it is challenging to learn variations of novel classes from only a few training videos. Moreover, if we are interested in predicting both base and novel classes, the imbalanced training set~($|\X_b| \gg |\X_n|$) will lead to a biased classifier that is dominated by base classes. To this end, we introduce a feature generator $G(z, \varphi(\y))$ that takes as inputs a Gaussian noise $z\in \mathbb{R}^{d_z}$ and a semantic embedding $\varphi(\y)$, and produces a CNN video feature $\phi^{\prime} \in \mathbb{R}^{d_x}$ of class $\y$. Here, the video feature space is defined by the 3D CNN $\phi$ which extracts clip-level features and is learned at the representation learning stage described in Section~\ref{sec:3dfsv}. Given a semantic embedding $\phi(\y)$, by repeatedly sampling noise $z$ and recomputing $\phi^{\prime}=G(z, \varphi(\y))$, arbitrarily many features $\phi^{\prime}$ can be generated. After repeating the feature generation process for every novel class, we obtain a set of generated features $\X_g=\{(\phi^{\prime}, \y)| \y \in C_n\}$. If the end task is to predict only novel classes, we learn a classifier by optimizing the cross-entropy loss defined in Equation~\ref{eq:ce} on the combination of generated features $\X_g$ and real features extracted from real videos $X_n$. In the more realistic setting where both base and novel classes are predicted, the training set becomes the union of generated features $\X_g$ and real features extracted from real videos $X_n$ and $X_b$. During the classifier learning, we only learn the classifier weights while the feature generator $G$ and feature extractor $\phi$ are fixed.         

%\matthijs{Relate to  \cite{xian2018feature} and \cite{zhu2018generative}}

\myparagraph{VFGAN training.} The goal of the generator $G$ is to synthesize video features that help to learn a better classifier. Intuitively, the generated features are useful when they mimic the distribution of real features. To this end, we adopt the GAN framework to learn the real feature distribution because of its strength to capture complex data distributions. More specifically, we introduce a discriminator $D(\phi(\x_i),\varphi(\y))$ that tries to distinguish real features $\phi(\x_i)$ and generated features $\phi^{\prime}$ conditioned on class embedding $\varphi(\y)$. Our feature generator is learned by optimizing the following Wasserstein GAN loss, 
\begin{align}
\label{eq:wgan}
\min_G \max_D \mathcal{L}_{WGAN} =& E[D(\phi(\x_i),\varphi(\y))] - E[D(\phi^{\prime},\varphi(\y))]  \\ 
                   & - \lambda E[(||\nabla_{\hat{\phi}} D(\hat{\phi},\varphi(\y))||_2 - 1)^2], \nonumber
\end{align}
where $\phi(\x_i)$ denotes the real feature of video clip $x_i$ extracted from the feature extractor $\phi$,  $\y$ is its class label,  $\phi^{\prime}=G(z, \varphi(\y))$ denotes the generated feature, $\hat{\phi} = \alpha \phi(\x_i) + (1 - \alpha)\phi^{\prime}$ with $\alpha \sim
U(0, 1)$, and $\lambda$ is the penalty coefficient. The first two terms approximate the Wasserstein distance between the real and generated distributions if $D$ is sufficiently smooth, while the third term enforces the smoothness of $D$ by penalizing its gradient with respect to the input. We chose the Wasserstein GAN loss because of its property of stable training~\cite{arjovsky2017towards}. Although it is known that GANs are hard to train for the image generation task~\cite{arjovsky2017towards}, we did not observe any instability issue in our framework because our model generates features where the dimensionality is much lower than the raw images. Unless otherwise stated, the Equation~\ref{eq:wgan} is optimized over base class videos $\X_b$, novel class videos $\X_n$ and the semantic embedding $\varphi$ e.g., word2vec or BERT. 
Note that video feature extractor $\phi$ and class embedding $\varphi$ are kept fixed during the feature generator training. As the retrieval method and VFGAN are complementary, we also propose to combine both methods by expanding the training set of the feature generator with the retrieved video clips $\X_p$. 
%\section{Generalized Many-Way Setting}

%\myparagraph{Combining video retrieval and VFGAN.} 

\section{Experiments}

%In this section, we first describe the datasets and implementation details. Then we discuss the limitation of prior few-shot video classification setup and introduce our more challenging evaluation setup called generalized many-way few-shot video classification setting.

In this section, we first provide the implementation details that reproduce our results. We then present the results comparing our approaches with the state-of-the-art methods on the existing Kinetics and SomethingV2 benchmarks. Finally, we show the results of our approaches in our proposed settings followed by model analysis.

\myparagraph{Datasets.} Kinetics~\cite{kay2017kinetics} is a large-scale video dataset which covers 400 human action classes including human-object and human-human interactions. Its videos are collected from Youtube and trimmed to include only one action class. The UCF101~\cite{soomro2012ucf101} dataset is also collected from Youtube videos, consisting of 101 realistic human action classes, with one action label in each video. SomethingV2~\cite{goyal2017something} is a fine-grained human action recognition dataset, containing 174 action classes, in which each video shows a human performing a predefined basic action, such as ``picking something up'' and ``pulling something from left to right''. We use the second release of the dataset. YFCC100M~\cite{thomee2015yfcc100m} is the largest publicly available multimedia collection with about 99.2 million images and 800k videos from Flickr. Although none of these videos are annotated with a class label, half of them~(400k) have at least one user tag. We use the tag-labeled videos of YFCC100M to improve the few-shot video classification.

\myparagraph{Data Splits.} Kinetics, UCF101 and SomethingV2 are used as our benchmark datasets with disjoint sets of base, validation and novel classes (see Table~\ref{tab:dataset} for details). 
%Here we refer to base classes as train classes. Test classes include the classes we sample novel classes from in each testing episode.
For Kinetics and SomethingV2, we follow the splits proposed by \cite{zhu2018compound} and \cite{cao2019few} respectively for a fair comparison. It is worth noting that 3 out of 24 test classes in Kinetics appear in Sports1M, which is used for pretraining our 3D ConvNet. But the performance drop is negligible if we replace those 3 classes with other 3 random kinetics classes that are not present in Sports1M. In addition, we create a new data split for UCF101 dataset, which is widely used for benchmarking video action classification as well. Following the same convention, we randomly select 64, 12 and 24 non-overlapping classes as base, validation and novel classes from the UCF101 dataset. We ensure that in our splits the novel classes do not overlap with the classes of Sports1M. For the generalized few-shot video classification setting, in each dataset the test set includes both samples from base classes coming from the validation split of the original dataset and samples belonging to  novel classes. 

{
\setlength{\tabcolsep}{4pt}
\renewcommand{\arraystretch}{1.2}
\begin{table}[t]
 \centering
 %\resizebox{\linewidth}{!}{%
   \begin{tabular}{l  c c c |c c c }
     & \multicolumn{3}{c|}{\textbf{\# classes}} & \multicolumn{3}{c}{\textbf{\# videos}} \\
     & \textbf{base}  & \textbf{val} &  \textbf{novel} & \textbf{train} & \textbf{val} & \textbf{test}   \\
     \hline
    \textbf{Kinetics} & 64 & 12 & 24 & 6400  & 1200  & 2400+2288 \\ 
    \textbf{UCF101} & 64 & 12 & 24  & 5891  & 443  & 971+1162 \\
    \textbf{SomethingV2} & 64 & 12 & 24  & 67013  & 1926 & 2857+5243
    %\vspace{1mm}
        % \hline
\end{tabular}
%}
\caption{Statistics of our data splits on Kinetics, UCF101 and SomethingV2 datasets. We follow the base, val, and novel class splits of \cite{zhu2018compound} and \cite{cao2019few} on Kinetics and SomethingV2 respectively. In addition, we add test videos~(the second number under the second test column) from base classes for our proposed benchmarks. We also introduce a new data split on UCF101.}
\label{tab:dataset}
\end{table}
}

\myparagraph{Implementation details.} %At training time, we use base classes to learn representations and validation classes to tune hyperparameters. 
Unless otherwise stated, our backbone is a 34-layer R(2+1)D~\cite{tran2018closer} pretrained on Sports1M~\cite{karpathy2014large}. It takes as input a video clip consisting of $F=16$ RGB frames with spatial resolution of $H=112\times W=112$. We extract clip features from the $d_v=512$ dimensional top pooling units of the R(2+1)D.  In the representation learning stage, we fine-tune the R(2+1)D with the Adam optimizer and a constant learning rate $0.001$ on all datasets and stop training when the validation accuracy of base classes saturates. We perform standard spatial data augmentation including random cropping and horizontal flipping. We also apply temporal data augmentation by randomly drawing 8 clips~(16 frames) from a video in one epoch. 
In the few-shot learning stage, the same data augmentation is applied and the novel class classifier is learned with a constant learning rate $0.01$ for $10$ epochs on all the datasets. At test time, we randomly draw $L=10$ clips from each video and average their predictions to produce the video-level prediction. We follow the episodic evaluation protocol~\cite{zhu2018compound} and report the top-1 accuracy averaged over 500 episodes. 

As for the retrieval step, we use the 300-dim~($d_t=300$) fasttext~\cite{joulin2016fasttext} embedding pretrained with GoogleNews. We first retrieve $N=20$ candidate videos for each class with video tag retrieval from the YFCC100M dataset~\cite{thomee2015yfcc100m}. Then we randomly draw 15 clips from each candidate videos and select $M=5$ best clips among those clips with visual similarities. We adopt FAISS~\cite{JDH17} for the k-nearest neighbor search with cosine similarity.  

We implement both the generator and discriminator of our VFGAN with the multilayer perceptron~(MLP). The generator consists of an input layer which concatenates the random noise $z$ and semantic embedding $\varphi(\y)$, a FC layer with 4096 hidden units followed by the LeakyReLU activation function, and an output FC layer with the Sigmoid function that produces a 512-dim feature vector. The discriminator also has only one hidden layer with 4096 hidden units. Its input layer concatenates the semantic embedding $\varphi(\y)$ and a 512-dim input feature~(generated or real), while the output layer produces a scalar without using any activation function~\cite{arjovsky2017wasserstein}. 
%We did not observe any performance boost if deeper neural networks are used. 
The range of real features are rescaled to be in [0, 1] because it is beneficial for the convergence. As for the semantic embedding, we use 300-dim fasttext embeddings of the class names for Kinetics and UCF101 datasets. On SomethingV2, every class name is described by a sentence and therefore we extract $768$-dim sentence embeddings from the pretrained BERT~\cite{reimers-2019-sentence-bert} model. The random noise $z$ is drawn from a unit  Gaussian  with the same  dimensionality as the semantic embedding. We use $\lambda=10$ as suggested in WGAN~\cite{gulrajani2017improved} and $\beta=0.01$ across all the datasets. 

\subsection{Comparing with SOTA on existing benchmark}
\label{sec:old_setting}

In this section, we compare our model with SOTA (state-of-the-art) on existing benchmarks which mainly consider 1-shot, 5-way and 5-shot, 5-way problems and evaluate only on novel classes. 

{
\setlength{\tabcolsep}{5pt}
\renewcommand{\arraystretch}{1.1}
\begin{table}[t]
 \centering
   \begin{tabular}{l c c  c  c  c}
     & & \multicolumn{2}{c}{\textbf{Kinetics}} & \multicolumn{2}{c}{\textbf{SomethingV2}} \\
     \hline
    \textbf{Method} &   \textbf{Backbone} & \textbf{1-shot} & \textbf{5-shot} & \textbf{1-shot} & \textbf{5-shot}  \\
    %\hline
%    Matching Net~\cite{zhu2018compound} & 53.3  & 74.6 & - & - \\ 
%    MAML~\cite{zhu2018compound} & 54.2 & 75.3 & - & -\\ 
    CMN~\cite{zhu2018compound} & ResNet & 60.5  & 78.9 & - & - \\ 
%    TSN++~\cite{cao2019few} & 64.5 & 77.9 & 33.6  & 43.0\\ 
    CMN++~\cite{cao2019few} & ResNet &  65.4 & 78.8 & 34.4  &  43.8\\ 
%    TRN++~\cite{cao2019few} & 68.4 & 82.0 & 38.6 & 48.9\\ 
    TAM~\cite{cao2019few} & ResNet &  73.0 & 85.8 & 42.8  & 52.3 \\
    TARN~\cite{bishay2019tarn} & C3D & 66.6 & 80.7 & - & - \\
    %    \hline
    %3DFSV~(ours, scratch) &  48.9 & 67.8  & 57.9 &  75.0 \\
   % 3DFSV~(Ours-selfPre) & 51.9 & 72.5 &  &   \\
   \hline
    TSL & R(2+1)D  & 92.5 & 97.8 & 59.1 & 80.1  \\
    TSL w/ Retrieval &  R(2+1)D &  95.3 & 97.8 & - & -  \\
    TSL w/ VFGAN &  R(2+1)D &  94.9 & \textbf{98.4} & \textbf{66.7} & \textbf{82.2}  \\
    TSL w/ both &  R(2+1)D & \textbf{96.2}  & \textbf{98.4}  & - & -  \\
    %\vspace{1mm}
    %\hline
    \end{tabular}
\caption{Comparing with the state-of-the-art few-shot video classification methods. We report top-1 accuracy on the novel classes of Kinetics and SomethingV2 for 1-shot and 5-shot tasks~(both in 5-way). TSL: our proposed two-stage learning baseline. TSL w/ Retrieval: TSL with retrieved videos. TSL w/ VFGAN: TSL with generated features. TSL w/ both: TSL with both generated features and retrieved videos. CMN~\cite{zhu2018compound} and TARN~\cite{bishay2019tarn} did not report results on the SomethingV2 dataset.
%Note that CMN~\cite{zhu2018compound} did not report results on SomethingV2 and we do not retrieve videos for SomethingV2 because it is unlikely that YFCC100M~\cite{thomee2015yfcc100m} includes any relevant instructional videos. 
}
\label{tab:main}
\end{table}
}

\myparagraph{Existing few-shot video classification benchmarks.} The existing practice of~\cite{zhu2018compound} and ~\cite{cao2019few} indicates randomly selecting 100 classes on Kinetics and on SomethingV2 datasets respectively. Those 100 classes are then randomly divided into 64, 12, and 24 non-overlapping classes to construct the meta-training, meta-validation and meta-testing sets. 
The meta-training and meta-validation sets are used for training models and tuning hyperparameters. In the testing phase of this meta-learning setting~\cite{zhu2018compound,cao2019few}, each episode simulates a $n$-way, $k$-shot classification problem by randomly sampling a support set consisting of $k$ samples from each of the $n$ classes, and a query set consisting of one sample from each of the $n$ classes. While the support set is used to adapt the model to recognize novel classes, the classification accuracy is computed at each episode on the query set and mean top-1 accuracy over 20,000 episodes constitutes the final accuracy.

\myparagraph{Baselines.} The baselines CMN~\cite{zhu2018compound} and TAM~\cite{cao2019few} are considered as the state-of-the-art in few-shot video classification. CMN~\cite{zhu2018compound} proposes a multi-saliency embedding function to extract video descriptor, and few-shot classification is then done by the compound memory network~\cite{kaiser2017learning}. %CMN also provides few-shot video classification results for MAML and MatchNet, which are popular in few-shot image classification. 
TAM~\cite{cao2019few} proposes to leverage the long-range temporal ordering information in video data through temporal alignment. They additionally build a stronger CMN, namely CMN++,  by using the few-shot learning practices from \cite{chen2019closer}. TARN~\cite{bishay2019tarn} proposes to utilize attention mechanisms to perform temporal alignment.  We use their reported numbers for fair comparison. The results are shown in Table~\ref{tab:main}.

\myparagraph{Results.} On both Kinetics and SomethingV2 datasets with different numbers of training shots, our TSL baseline, i.e. without retrieval and feature generation, outperforms the previous best methods by a wide margin. In particular, on Kinetics, TSL improves TAM by over $19\%$ in 1-shot case~($73.0\%$ of TAM vs $92.5\%$ of TSL). On SomethingV2, TSL again boosts the performance of TAM by $15.1\%$ in 1-shot~($42.8\%$ of TAM vs $57.9\%$ of TSL) and by surprisingly $22.7\%$ in 5-shot~($52.3\%$ of TAM vs $75.0\%$ of TSL). Note that TAM extracts image-level features from the ResNet~\cite{he2016deep} while our approach extracts clip-level features with the recent R(2+1)D~\cite{tran2018closer}. Those results show that a simple linear classifier on top of a 3D CNN, e.g.  R(2+1)D~\cite{tran2018closer}, performs better than sophisticated methods with a 2D ConvNet as a backbone. This suggests that we should use modern video representation network like R(2+1)D instead of 2D ConvNet for the few-shot video classification task because the temporal information is important.

Although as shown in C3D~\cite{tran2015learning}, I3D~\cite{carreira2017quo}, R(2+1)D~\cite{tran2018closer}, spatiotemporal CNNs have an edge over 2D spatial ConvNet~\cite{he2016deep} in the fully supervised video classification with enough annotated training data, we are the first to show the big performance boost of R(2+1)D for the few-shot video classification task. Recently, TARN~\cite{bishay2019tarn} proposes to adopt the C3D~\cite{tran2015learning} pretrained on Sports1M as the backbone. Specifically, TARN~\cite{bishay2019tarn} skips the representation learning stage and directly extracts features from the fixed C3D followed by their few-shot learning algorithm. In contrast, we acknowledge that there is a substantial domain gap between the pretraining and target datasets and perform end-to-end finetuning of the 3D CNN on the base classes. As shown in Table~\ref{tab:main}, our TSL significantly outperforms TARN~\cite{bishay2019tarn} in terms of few-shot learning accuracy on Kinetics~(5-way 1-shot: 66.6\% of TARN vs 92.5\% of TSL, 5-way 5-shot: 80.7\% of TARN vs 97.8\% of TSL). These results again confirm the effectiveness of our two-stage baseline.

%Our results in Table 2
%\zeynep{are these results in the paper? if so, refer to it, if not indicate that this is an additional information. it is not clear here.} 
%show that 3D CNNs trained from random initialization significantly outperform pretrained 2D CNNs on SomethingV2 dataset which requires strong temporal information. This confirms the importance of 3D CNNs for few-shot video classification. 

\begin{figure*}[t]
	\centering
	\includegraphics[width=.29\linewidth, trim=10 0 65 0,clip]{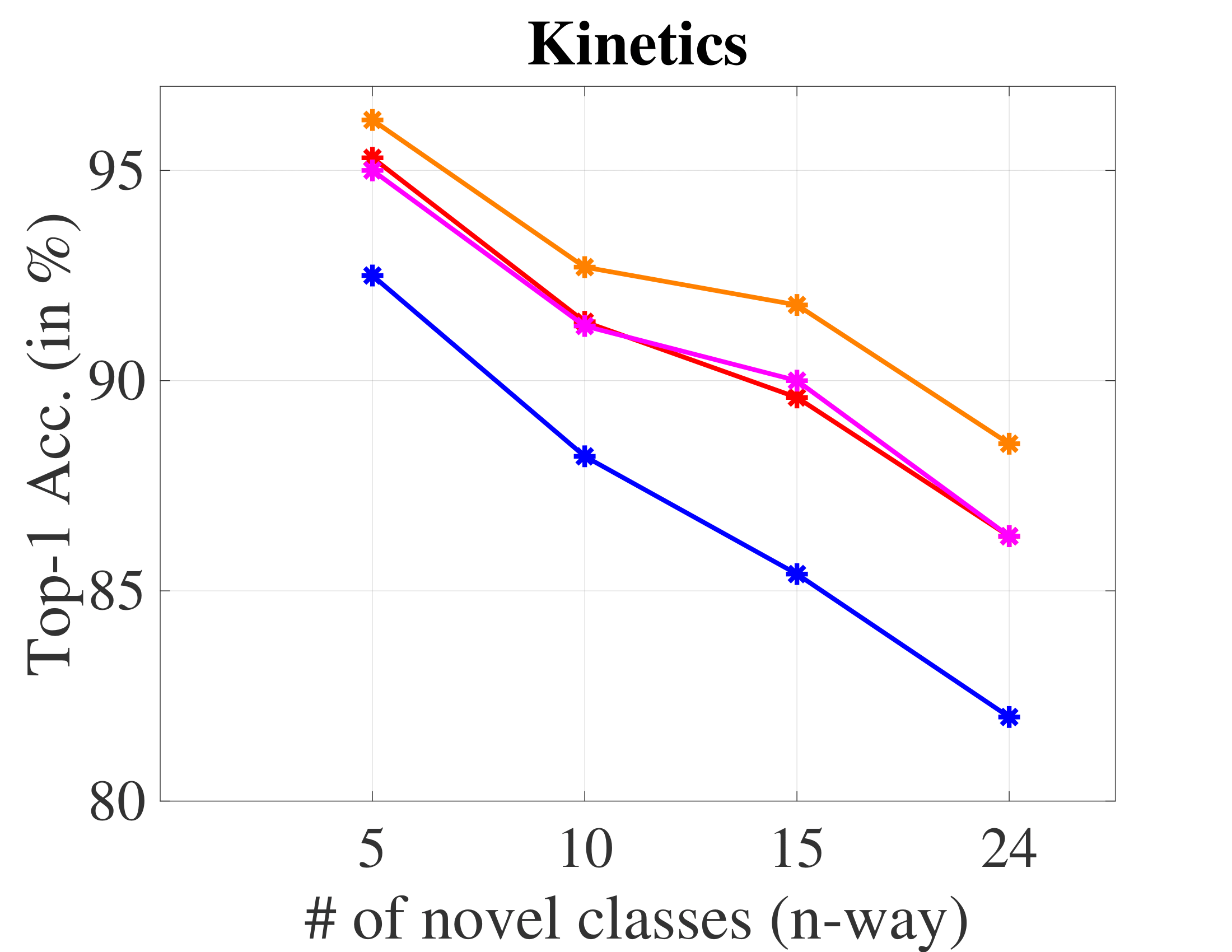}
	\includegraphics[width=.29\linewidth, trim=10 0 65 0,clip]{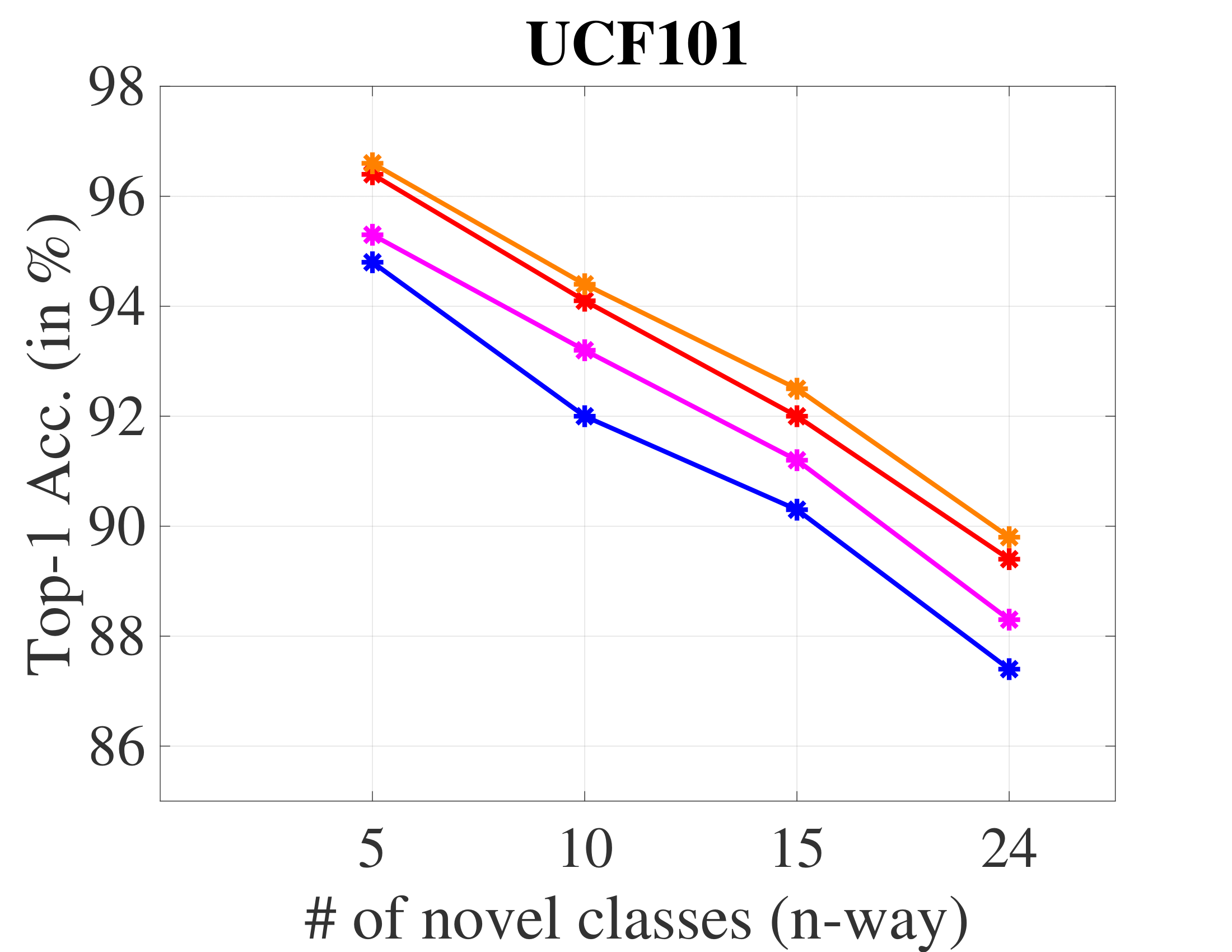} 
	\includegraphics[width=.29\linewidth, trim=10 0 65 0,clip]{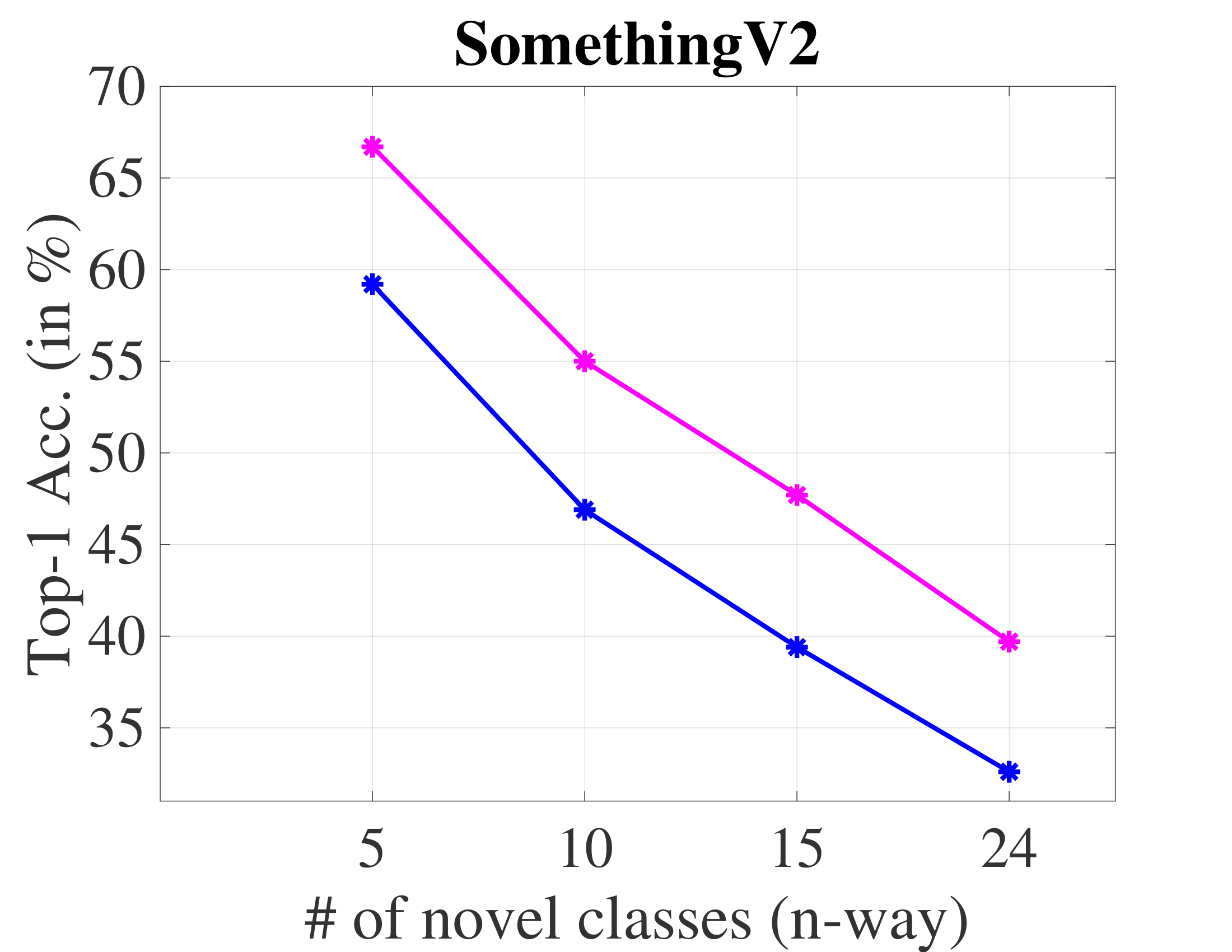}
	\includegraphics[width=.1\linewidth, trim=500 0 50 0,clip]{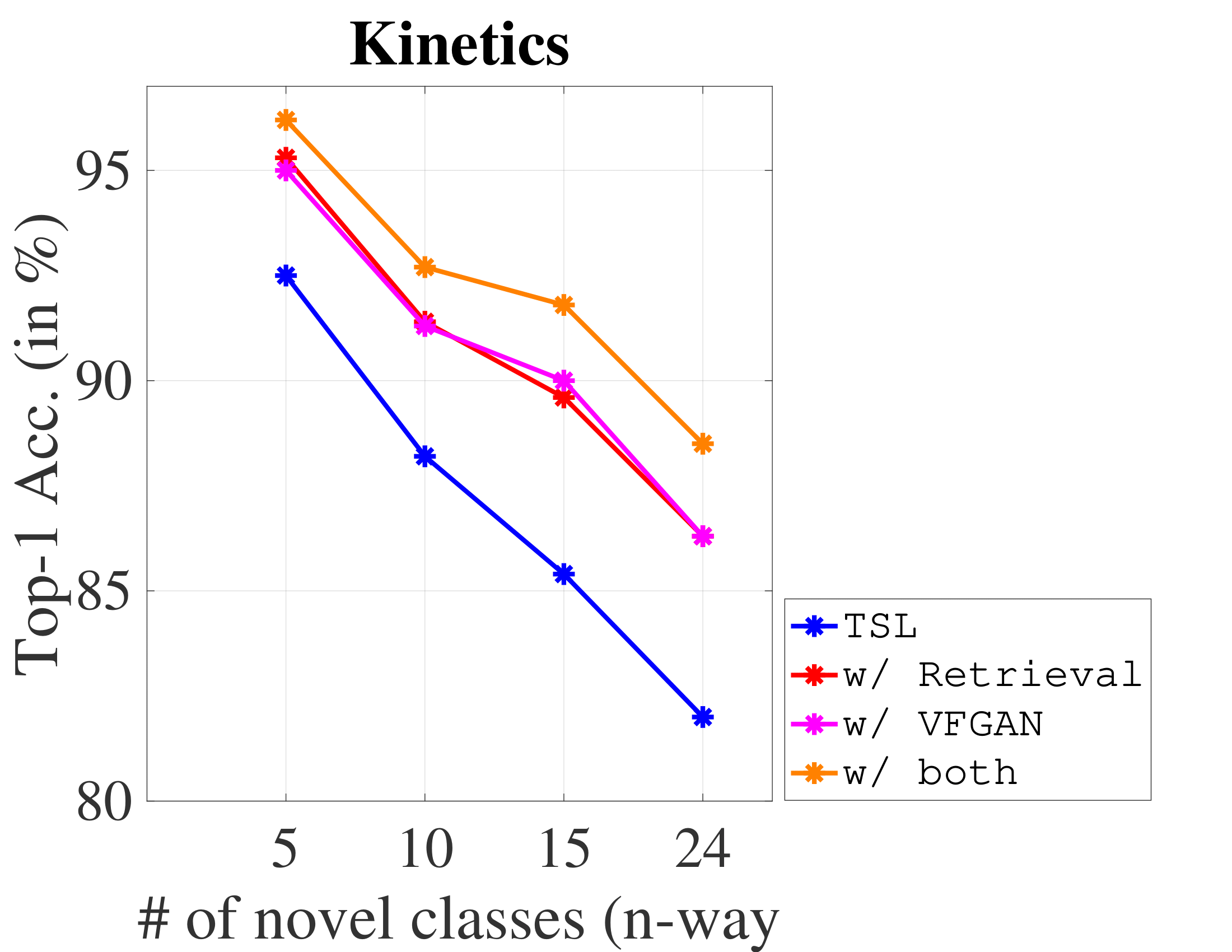} \\
	\caption{Results of our TSL baseline, our TSL with retrieval~(w/ retrieval) , with feature generation~(w/ VFGAN) and with both techniques~(w/ both) on Kinetics, UCF101 and SomethingV2 datasets in the many-way 1-shot video classification setting. More specifically,  we go beyond the prior 5-way classification setting and evaluate 5, 10, 15 and 24 (all) of the novel classes in each testing episode. We report the top-1 accuracy of novel classes.}
	\label{fig:more_way}
\end{figure*}

Our TSL w/ Retrieval, i.e. with retrieved tag-labeled video clips, lead to further improvements in 1-shot case~(TSL $92.5\%$ vs TSL w/ Retrieval $95.3\%$) on Kinetics dataset. This implies that tag-labeled videos retrieved from the YFCC100M dataset~\cite{thomee2015yfcc100m} include discriminative cues for novel classes.  In 5-shot, TSL w/ Retrieval achieves similar performance as our TSL approach because this task is almost saturated with an accuracy of $97.8\%$. It is worth noting that our retrieval algorithm is highly efficient such that it scales to utilize 8M tag-labeled videos. 
We do not retrieve any tag-labeled videos for the SomethingV2 dataset because it is a fine-grained dataset of instructional actions and it is unlikely that YFCC100M includes any relevant video for that dataset. 

Moreover, we observe that our TSL w/ VFGAN outperforms the TSL baseline on all the datasets consistently. In particular, on SomethingV2, TSL w/ VFGAN improves TSL by over $7\%$ in 1-shot~(66.7\% of TSL w/ VFGAN vs 59.1\% of TSL). This indicates that our VFGAN is able to generate discriminative video features that are complementary to the training data. Besides, TSL w/ VFGAN achieves comparable results with Ours w/ Retrieval. This is encouraging because learning the VFGAN does not need any additional videos while TSL w/ Retrieval requires a large collection of tag-labeled videos that are not always available. The semantic embeddings used by VFGAN are easy to obtain from pretrained language models like Fasttext~\cite{joulin2016fasttext} or BERT~\cite{devlin2018bert}. Finally, we show that combining our VFGAN and retrieval method~(i.e., TSL w/ both) outperforms applying them separately in 1-shot 5-way setting on Kinetics, indicating that VFGAN and video retrieval are complementary.
As a summary, although the prior 5-way classification setting is still challenging to those methods with 2D ConvNet backbone, we argue that the results saturate with the stronger spatiotemporal CNN backbone on Kinetics dataset.

\subsection{Many-way few-shot video classification}
\label{sec:manyway_setting}

The previous evaluation protocol only considers 5 novel classes randomly drew from 24 novel classes in each testing episode i.e., a 5-way classification problem. Although such setting remains challenging for previous methods~\cite{zhu2018compound, cao2019few} with a 2D CNN backbone, the performance of our approaches with a 3D CNN backbone is nearly saturated i.e., TSL w/ VFGAN obtains a top-1 accuracy of $98.4\%$ in 5-shot on Kinetics dataset~(Table.~\ref{tab:main}). In this section, we introduce the many-way few-shot video classification setting where we extend the 5-way classification to 10-way, 15-way and 24-way settings. For example, in 1-shot 10-way setting, each testing episode includes 10 novel classes which are randomly drew from 24 novel classes. Then we sample randomly a support set consisting of 1 training video from each of the 10 classes and a query set consisting of 15 testing videos from each of the same classes. The top-1 classification accuracy is computed at each episode on the query set of the 10 classes and we report the mean accuracy over 500 episodes. We then present and discuss the results of our three approaches in the many-way few-shot video classification settings on Kinetics, SomethingV2 and UCF101 datasets. As the CMN~\cite{zhu2018compound} and TAM~\cite{cao2019few} codes were not available at submission we do not include their results here.

\myparagraph{Results.} Figure~\ref{fig:more_way} shows the results of many-way few-shot video classification where we extend the number of novel classes at test time to 10, 15 and 24 classes. Our TSL w/ retrieval, TSL w/ VFGAN and TSL w/ both achieve over 95\% accuracies in both Kinetics and UCF101 datasets for 5-way classification. With the increasing number of novel classes, e.g. 10, 15 and 24, as expected, the performance monotonically degrades. We observe that our approaches with retrieval and VFGAN consistently outperforms our baseline TSL across different number of novel classes and datasets. The best performance is achieved by combining VFGAN and retrieval~(i.e., TSL w/ both). The more challenging the task becomes, e.g. from 5-way to 24-way, the larger improvement retrieval and feature generation approaches can achieve on Kinetics, i.e. our retrieval-based method is better than our baseline TSL by $2.8\%$ in 5-way~( TSL $92.5\%$ vs TSL w/ retrieval $95.3\%$ ) and the gap becomes $4.3\%$ in 24-way~( TSL  $82.0\%$  vs TSL w/ retrieval $86.3\%$). On the one hand, the improvement confirms that both retrieving tag-labeled videos and generating video features from semantic embeddings are effective approaches to tackle many-way few-shot video classification. On the other hand, the trend with a decreasing accuracy by going from 5-way to 24-way indicates that the more realistic task on few-shot video classification has not yet been solved even with the current best video feature backbone i.e., R(2+1)D. In particular, on SomethingV2 dataset, we found that our approach with feature generation only obtains an accuracy of $40.0\%$, having a big gap with the performance on Kinetics and UCF101.  We hope that these results will encourage more progress in this challenging setting of many-way few-shot video classification setting.

{
\setlength{\tabcolsep}{2pt}
\renewcommand{\arraystretch}{1.2}
\begin{table}[t]
 \centering
 \resizebox{\linewidth}{!}{%
  \begin{tabular}{c l | c c c | c c c| c c c}
 &  & \multicolumn{3}{c}{\textbf{Kinetics}} & \multicolumn{3}{c}{\textbf{UCF101}} & \multicolumn{3}{c}{\textbf{SomethingV2}}  \\ 
  & \textbf{Method} & novel & base & HM & novel & base & HM  & novel & base  & HM \\ \hline
  \multirow{3}{*}{1-shot} & TSL & 7.5 & 88.7 & 13.8 & 3.5 & 97.1 & 6.8 & 0.0 & 59.5 & 0.0 \\
   & TSL w/ Retrieval & 13.7 & 88.7 & 23.7 & 4.9 & 97.1 & 9.3 & - & - & -\\
     %& Ours w/ VFGAN & 61.1 & 91.6 & 73.3 & 72.4 & 96.0 & 82.5 & 20.4 & 57.9 & 30.2\\
    5-way  & TSL w/ VFGAN & 81.7 & 91.0 & 86.1 & 87.6 & 94.6 &  91.0 & 20.4 & 57.9 & \textbf{30.2} \\
  & TSL w/ both &  86.3 & 90.9  & \textbf{88.5} & 89.1  & 94.0  & \textbf{91.5}  & - & - & - \\
     \hline
   \multirow{3}{*}{5-shot} & TSL & 20.5 & 88.7 & 33.3 & 10.1 & 97.1 & 18.3 & 0.0 & 59.5 & 0.0\\
    & TSL w/ Retrieval & 22.3 & 88.7 & 35.6  & 10.4 & 97.1 & 18.8 & - & - & -\\
     %& Ours w/ VFGAN & 80.6 & 91.6 & 85.7 & 84.9 & 96.0 & 90.1 & 31.9 & 58.3 & 41.2\\
       5-way & TSL w/ VFGAN & 94.3 & 90.6 & \textbf{92.4} & 95.2 & 93.9 & \textbf{94.5} & 31.9 & 58.3 & \textbf{41.2} \\
     & TSL w/ both & 94.2 & 90.7 & \textbf{92.4} & 95.1  & 93.9 & \textbf{94.5} & - & - & - \\
     \hline
     \end{tabular}
    }
\caption{Generalized few-shot video classification with our TSL baseline, our TSL with retrieval~(TSL w/ retrieval),  feature generation~(TSL w/ VFGAN) and both~(TSL w/ both) on Kinetics, UCF101 and SomethingV2 in the 5-way tasks. We report top-1 accuracy on base, novel classes and their harmonic mean~(HM). }
\label{tab:gfsv}
\end{table}
}

\subsection{Generalized few-shot video classification}
\label{sec:new_setting}
%\zeynep{the many way few-shot setting and the experiments seem to be presented completely separately from the results in generalized few-shot video classification. Suggestion: I would just present them in two different subsections grouped as (4.2 many way few-shot classification) and (4.3 eneralized few-shot video classification). How about that?}

%In the following paragraph, we evaluate our methods in the more realistic and challenging generalized few-shot video classification~(GFSV) setting.

Another limitation of the previous evaluation protocols is that they only report the performance on novel classes at the test time. This ignores the potential imbalance issues between base and novel classes and is based on the unrealistic assumption that all test videos come from novel classes. 
To address this issue, we propose a more challenging and realistic setting, namely generalized few-shot video classification~(GFSV), where the model needs to predict both base and novel classes. To evaluate a $n$-way $k$-shot problem in GFSV, in addition to a support and a query set of novel classes, at each test episode we randomly draw an additional query set of 15 samples from each of the 64 base classes. We do not sample a support set for base classes because base class classifiers have been learned during the representation learning phase~(See Equation~\ref{eq:rl}). As for the evaluation metric, we compute the top-1 accuracies on base classes, novel classes, and the harmonic mean, which is defined as $HM = \frac{2 * Acc_{base} * Acc_{novel}}{Acc_{base} + Acc_{novel}}$
%\begin{equation}
%    HM = \frac{2 * Acc_{base} * Acc_{novel}}{Acc_{base} + Acc_{novel}}
%\end{equation}
where $Acc_{base}$ and $Acc_{novel}$ denote the base accuracy and novel accuracy respectively. We chose the harmonic mean because it measures how well the model can balance base and novel class performance i.e., it is high if and only if both base and novel classes are high. Finally, we report the averaged base accuracy, mean novel class accuracy and harmonic mean over 500 episodes.

\myparagraph{Results.} We present the GFSV results in Table~\ref{tab:gfsv}. Across all the three datasets, we observe a substantial performance gap between base and novel classes for our TSL baseline and retrieval-based approach. For instance, on the Kinetics dataset,  TSL only achieves $7.5\%$ on novel classes vs $88.7\%$ on base classes, leading to a low harmonic mean of $13.8\%$. 
%on the SomethingV2 dataset, the novel class accuracy of our TSL is nearly zero while base classes obtains an accuracy of $59.5\%$, leading to a zero harmonic mean. 
The reason is that predictions of novel classes are dominated by the base classes due to the imbalance issues. Interestingly, our video retrieval step~(TSL w/ retrieval) is able to improve the TSL baseline in terms of harmonic mean~(HM) e.g., $13.8\%$ of TSL vs $23.7\%$ of TSL w/ retrieval on Kinetics.  A similar trend can be observed on the UCF101 dataset. Those results demonstrate that our video step can alleviate the imbalance issues to some extent. Furthermore, we observe that our approach with feature generation~(ours w/ VFGAN) significantly improves the accuracy on novel classes with better performance or a slight performance drop on base classes, yielding the best harmonic mean on all the three datasets. For example, 
%ours w/ VFGAN improves our baseline by over $53\%$ on novel classes and over $2\%$ on base classes. 
on the challenging SomethingV2 dataset, TSL w/ VFGAN achieves a HM of $30.2\%$ vs $0.0\%$ of TSL in 1-shot, 5-way and a HM of $41.2\%$ vs $0.0\%$ of TSL in 5-shot, 5-way. Intuitively, our VFGAN generates additional video features for novel classes such that the training data becomes more balanced. Those results indicate that our VFGAN is able to generate video features of novel classes that resemble the real ones. Finally, we observe that combining both VFGAN and retrieval~(TSL w/ both) achieves the best HM in 1-shot 5-way on Kinetics, confirming again the complementarity of these two techniques. 

%Our TSL coupled with VFGAN is an effective way to tackle the imbalanced issue in the GFSV setting.% At the same time, we realize that generalized few-shot video classification (GFSV) setting, e.g. not restricting the test time search space only to novel classes but considering all of the classes even though base classes are distracting, is still a challenging task~(there is still big performance gap between novel and base classes) and hope that this setting will attract interest of a wider community for future research. 

%In contrast to the video retrieval, our VFGAN does not need any 

\myparagraph{Discussion.} The few-shot video classification setting~(FSV) has a strong assumption that test videos all come from novel classes. In contrast to the FSV, generalized few-shot video classification~(GFSV) is more realistic in the sense that it requires models to predict both base and novel classes in each testing episode. In other words, 64 base classes become distracting classes when predicting novel classes which makes the task more challenging. Distinguishing novel and base classes is difficult because there are severe imbalance issues between the base classes with a large number of training examples and the novel classes with only few-shot examples. We show that this imbalance issue can be partially addressed by augmenting the training data with external information i.e., tag-labeled videos or semantic embeddings of class names. Compared to the video retrieval approach, our VFGAN achieves a better performance and does not require any large-scale dataset. At the same time, we realize that the GFSV setting 
%e.g. not restricting the test time search space only to novel classes but considering all of the classes even though base classes are distracting, 
is still a challenging task~(there is still big performance gap between novel and base classes) and hope that this setting will attract interest of a wider community for future research.

%\begin{comment}

%\end{comment}

%\subsection{Model Analysis}
\subsection{Model Analysis: Video Retrieval}
In this section, we perform an ablation study to understand the importance of each component of our TSL baseline and TSL with retrieval. After the ablation study, we conduct experiments under different numbers of unlabeled video clips and evaluate the effect of the number of retrieved clips to the few-shot video classification (FSV) performance.   %\zeynep{I would make two separate sections to present (1) ablation study and effect of retrieved clips (2) training stability and the effect of number of generated video features using VGAN.}

{
\setlength{\tabcolsep}{5pt}
\renewcommand{\arraystretch}{1}
\begin{table}[t]
 \centering
   \begin{tabular}{c   c  c  c c | c | c | c}
     \textbf{PR}  & \textbf{RL} & \textbf{VR}  & \textbf{BD} & \textbf{BC} & \textbf{Kinetics100} & \textbf{UCF101} & \textbf{SomethingV2} \\
    \hline
     \checkmark   &  &  & & &  27.1 & 37.0 & 22.0  \\ %\hline
      &   \checkmark & &  & &  48.9  & 57.4 & 57.9 \\ %\hline
       %&  \checkmark & \checkmark & & & & 51.9   \\ %\hline
     \checkmark &   \checkmark & &  & & 92.5 & 94.8 & \textbf{59.1} \\ %\hline
     \checkmark &   \checkmark & \checkmark &  & & 91.4  & 94.1 & - \\ %\hline
    \checkmark &   \checkmark & \checkmark &  & \checkmark & 93.2 & 95.1 & -   \\ %\hline
    \checkmark &  \checkmark  & \checkmark & \checkmark & \checkmark & \textbf{95.3}  & \textbf{96.4} & -  \\
    \hline
    \end{tabular}
\caption{Ablation study on 5-way 1-shot video classification task on Kinetics, UCF101 and SomethingV2. \textbf{PR}: pretrain R(2+1)D on Sports1M;  \textbf{RL}: representation learning on base classes; \textbf{VR}: retrieve unlabeled videos with tags~\cite{thomee2015yfcc100m}; \textbf{BD}: batch denoising. \textbf{BC}: best clip selection. We report the top-1 accuracy on novel classes. }
\label{tab:ablation}
\end{table}
}

\myparagraph{Ablation study.} We ablate our model in the 1-shot, 5-way video classification task on Kinetics, UCF101 and SomethingV2 datasets with respect to five critical parts including pretraining R(2+1)D on Sports1M~(\textbf{PR}), representation learning on base classes~(\textbf{RL}), video retrieval with tags~(\textbf{VR}), batch denoising~(\textbf{BD}) and best clip selection~(\textbf{BC}). 
%In addition, we also made an attempt to adopt the self-supervised model of \cite{korbar2018cooperative} as the backbone~(\textbf{SS}) and show whether self-supervised learning is beneficial for few-shot video classification.  
%Table~\ref{tab:ablation} shows the results. 

We start from a model with only a few-shot learning stage on novel classes. If a \textbf{PR} component i.e., pretraining the R(2+1)D on Sports1M~\cite{karpathy2014large}, is added to the model~(first result row in Table~\ref{tab:ablation}), the newly-obtained model can achieve $27.1\%$ accuracy on Kinetics which is only slightly better than random guessing performance~($20\%$). It demonstrates that a pretrained 3D CNN alone is not sufficient for a good performance. Besides, it also indicates that there exists a domain shift between the pretraining dataset, i.e. Sports1M, and our target Kinetics dataset. 

On Kinetics dataset, adding \textbf{RL} component i.e., representation learning on base classes~(\textbf{RL}) , to the  model~(the second result row) means to train representation on base classes from scratch, which results in a worse accuracy of $48.9\%$ compared to our full model. The primary reason for worse results is that optimizing the massive number of parameters of R(2+1)D is difficult on a train set consisting of only 6400 videos. 
%Interestingly, if we adopt the self-supervised pretrained 3D CNN~(MC3 pretrained on Kinetics without using any label) of \cite{korbar2018cooperative}, i.e., \textbf{SS},  we immediate get $3.0\%$ performance gains~(the third result row) over training from random initialization.  
Adding both \textbf{PR} and \textbf{RL} components~(the fourth row) obtains an accuracy of $92.5$ which significantly boosts adding \textbf{PR} and \textbf{RL} components alone, indicating the importance of each stage in TSL.  However, on SomethingV2 dataset, even though we train the TSL from the random initialization , our results still remain promising~(only \textbf{RL} achieves $57.9\%$ vs TAM $42.8\%$ in Table~\ref{tab:main}). This confirms the importance of 3D CNNs for few-shot video classification.

Next, we study two critical components proposed in our retrieval step. 
%Finally, we study the effect of removing the best clip selection technique proposed in our retrieval approach. 
Comparing to our approach without retrieval~(the fourth row), directly appending tag-retrieved videos from YFCC100M~(\textbf{VR}) to the few-shot training set of novel classes~(the fifth result row) leads to $0.9\%$ performance drop, while performing the batch denoising~(the sixth row) in addition to \textbf{VR} obtains $0.7\%$ gain. This implies that noisy labels from retrieved videos may hurt the performance but our batch denoising technique handles the noise well. Finally, adding the best clip selection~(\textbf{BC}, the last row) after \textbf{VR} and \textbf{BD} gets a big boost of $2.8\%$ accuracy. This shows that filtering our the redundant clips in the candidate videos based on the visual similarities is critical. Similar trend can be observed from the results on UCF101. In summary, those ablation studies demonstrate the effectiveness of the five different critical parts in our approach.

\begin{figure}[t]
	\centering
		\includegraphics[width=.49\linewidth, trim=10 0 20 0,clip]{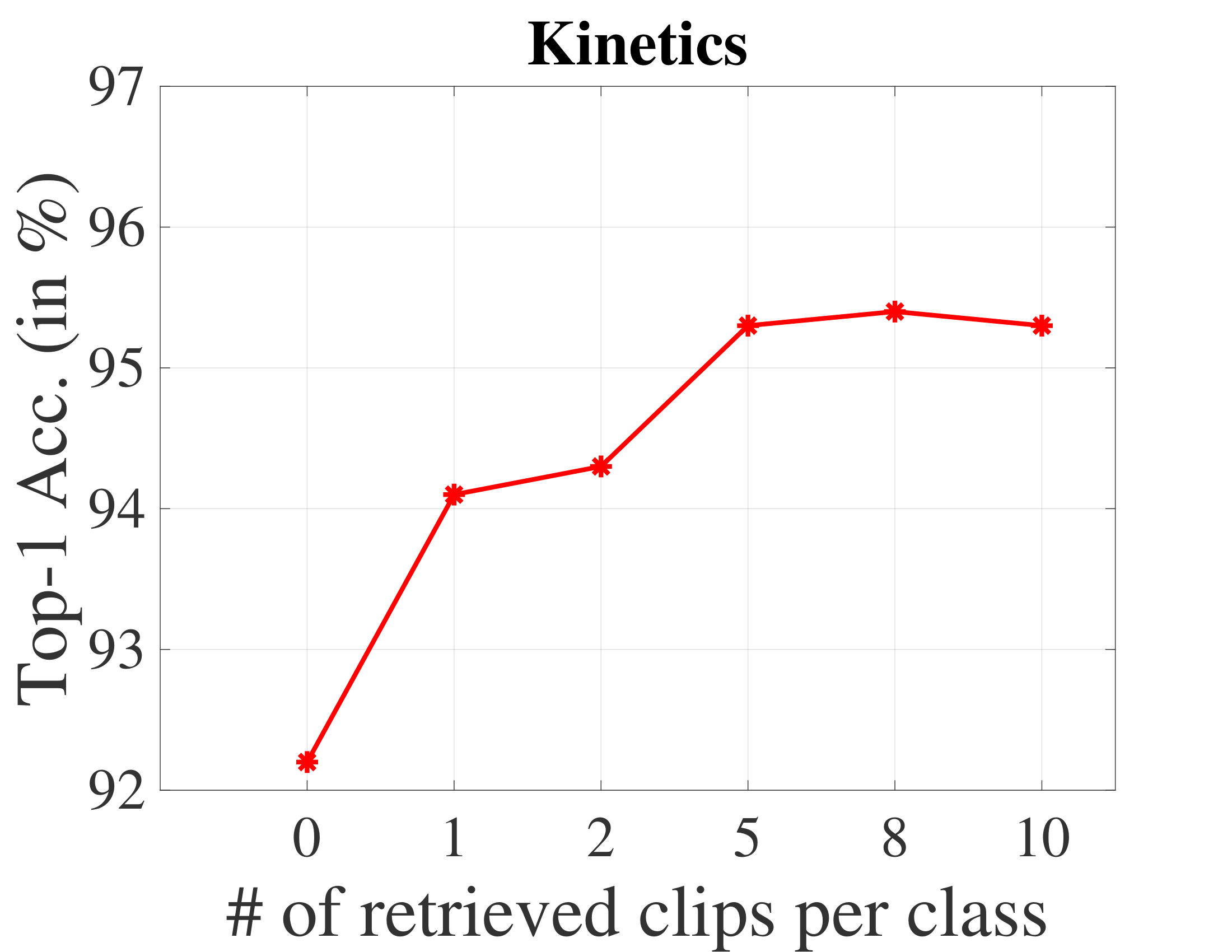}
		\includegraphics[width=.49\linewidth, trim=10 0 20 0,clip]{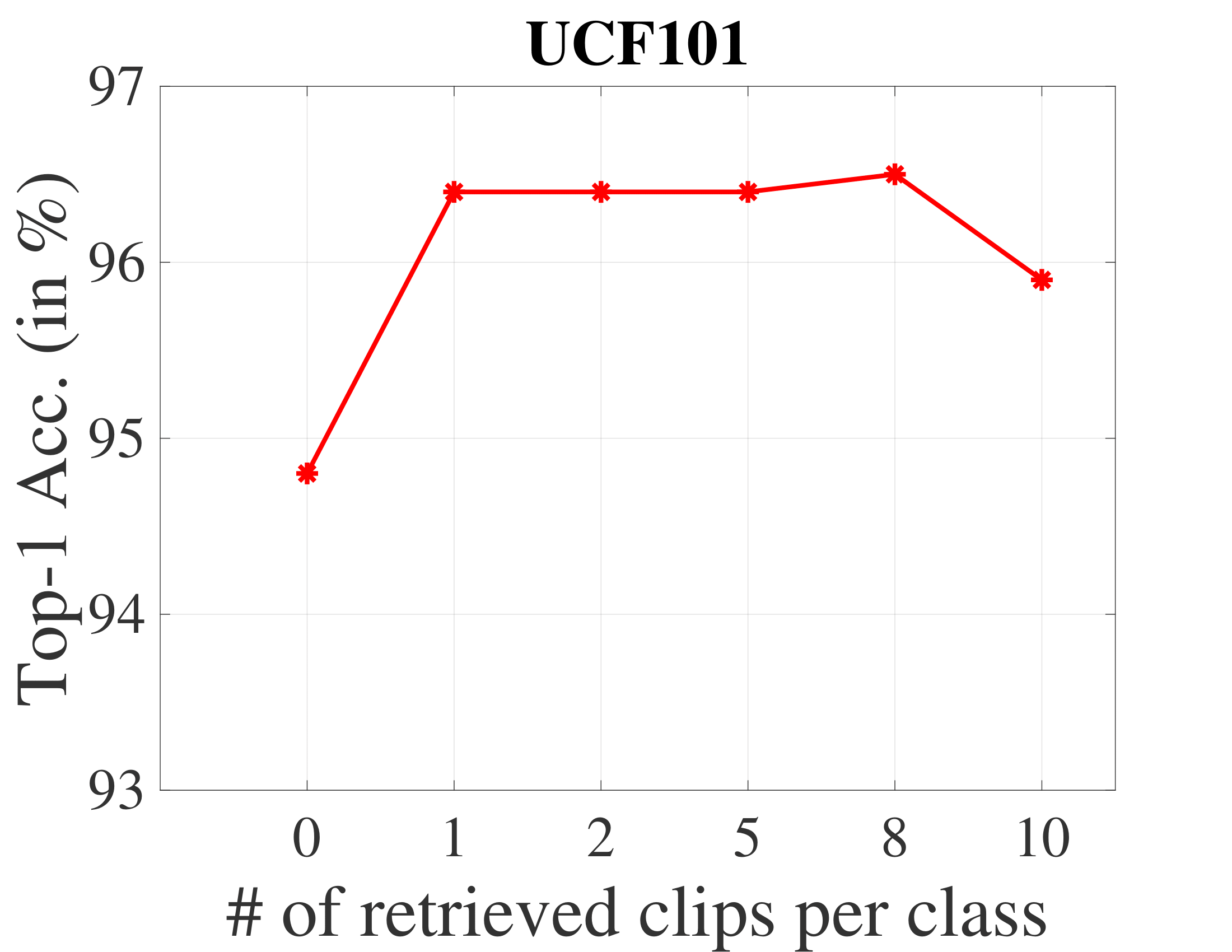}
	\caption{The effect of increasing the number of retrieved clips, \textbf{left}: on Kinetics, \textbf{right}: on UCF101. Both experiments are conducted on the 1-shot 5-way classification task, reporting top-1 accuracy in the few-shot video classification (FSV) setting.}
	\label{fig:num_retrieved_clips}
\end{figure}

\begin{figure}[t]
	\centering
		\includegraphics[width=.49\linewidth, trim=10 0 20 0,clip]{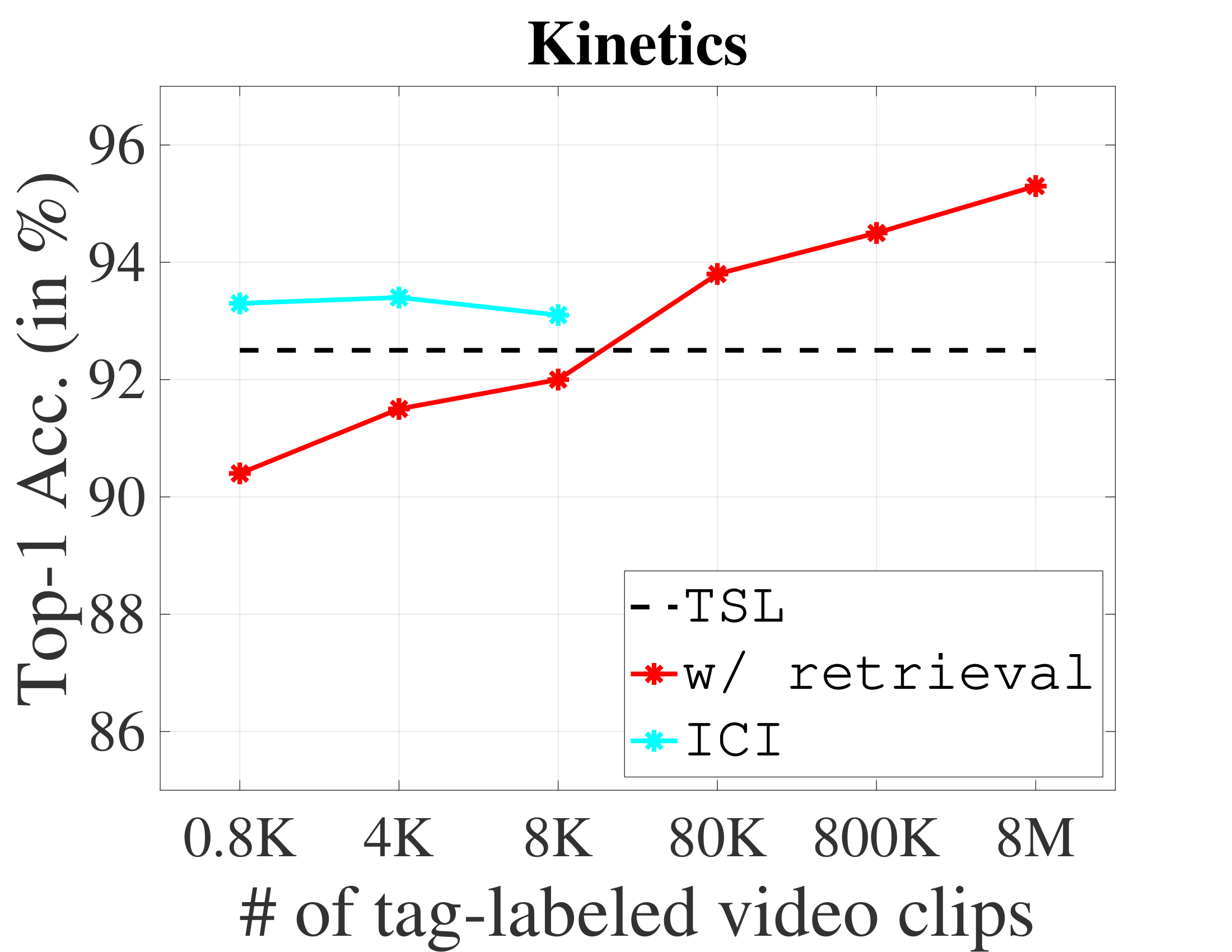}
		\includegraphics[width=.49\linewidth, trim=10 0 20 0,clip]{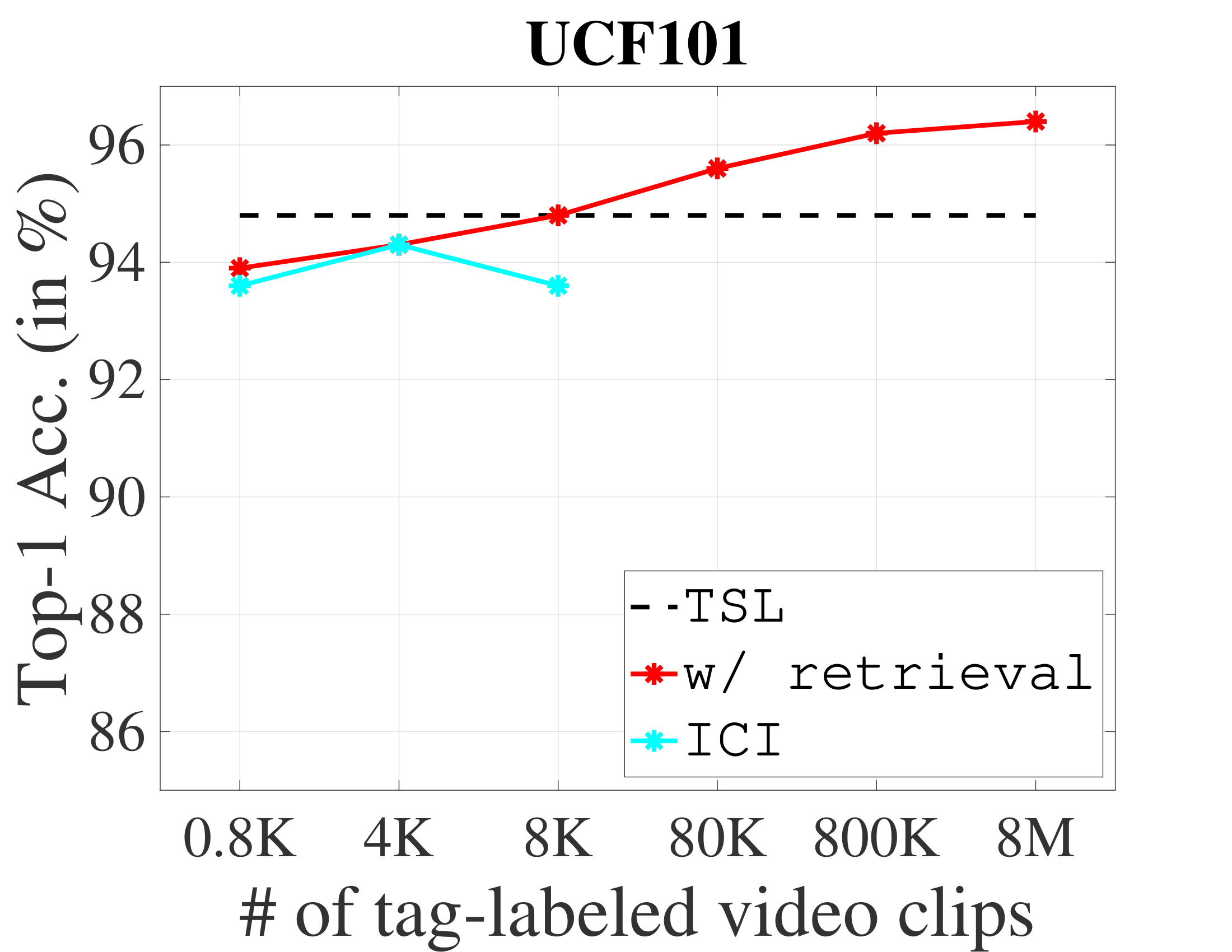}
	\caption{Comparing with a meta semi-supervised method i.e. ICI~\cite{wang2020instance} under different numbers of tag-labeled video clips, \textbf{left}: on Kinetics, \textbf{right}: on UCF101. Both experiments are conducted on the 1-shot 5-way few-shot video classification (FSV) setting.}
	\label{fig:num_unlabel}
\end{figure}

\myparagraph{Comparing with SOTA under different numbers of tag-labeled video clips.} ICI~\cite{wang2020instance} is the current SOTA in meta semi-supervised image classification. We adopt its released codes and adapt this method to address video classification by replacing its ResNet backbone with a R(2+1)D~\cite{tran2018closer} and tuning its hyperparameters for our task. In this comparison, we conduct experiments with increasing numbers of tag-labeled video clips by subsampling the YFCC100M dataset~\cite{thomee2015yfcc100m} i.e., from 0.8K to all 8M tag-labeled video clips, and show the results in Figure~\ref{fig:num_unlabel}. We have the following observations. First, we found that ICI does not scale to more than 80K tag-labeled examples because its pseudo-labeling algorithm is computationally expensive. Indeed, ICI maximally uses 400 unlabeled examples in its original paper~\cite{wang2020instance}. With a small number of tag-labeled examples~(i.e., 0.8K, 4K and 8K), we observe that both ICI and our TSL w/ retrieval do not show any advantage over the TSL baseline because those tag-labeled examples may not have sufficient diversity to cover the target novel classes. However, as we increase the number of tag-labeled examples, the performance of our TSL w/ retrieval keeps improving and reaches 95.3\% at 8M unlabeled examples on Kinetics and 96.4\% on UCF101. These results indicates that our retrieval approach is simple yet highly effective to generate pseudo-labels for novel classes. 

\myparagraph{Influence of the number of retrieved clips.}  
Intuitively, when the number of retrieved clips increases, the retrieved videos become more diverse, but at the same time, the risk of obtaining negative videos becomes higher. We show the effectiveness of our video retrieval step with the increasing number of retrieved clips in Figure~\ref{fig:num_retrieved_clips}. On the Kinetics dataset~(left of Figure~\ref{fig:num_retrieved_clips}), without retrieving any videos, the performance is $92.5\%$. As we increase the number of retrieved video clips for each novel class, the performance keeps improving and saturates at retrieving $8$ clips per class, reaching an accuracy of $95.4\%$. On the UCF101 dataset~(right of Figure~\ref{fig:num_retrieved_clips}), retrieving 1 clip gives us $1.6\%$ gain. Retrieving more clips does not further improve the results, indicating more negative videos are retrieved. We observe a slight performance drop at retrieving 10 clips because the noise level becomes too high, i.e. there are 10 times more noisy labels than clean labels.    
%On the other hand, our batch denoising strategy is able to tolerate the noise to some extent. 

\begin{figure}[t]
	\centering
	\includegraphics[width=.49\linewidth]{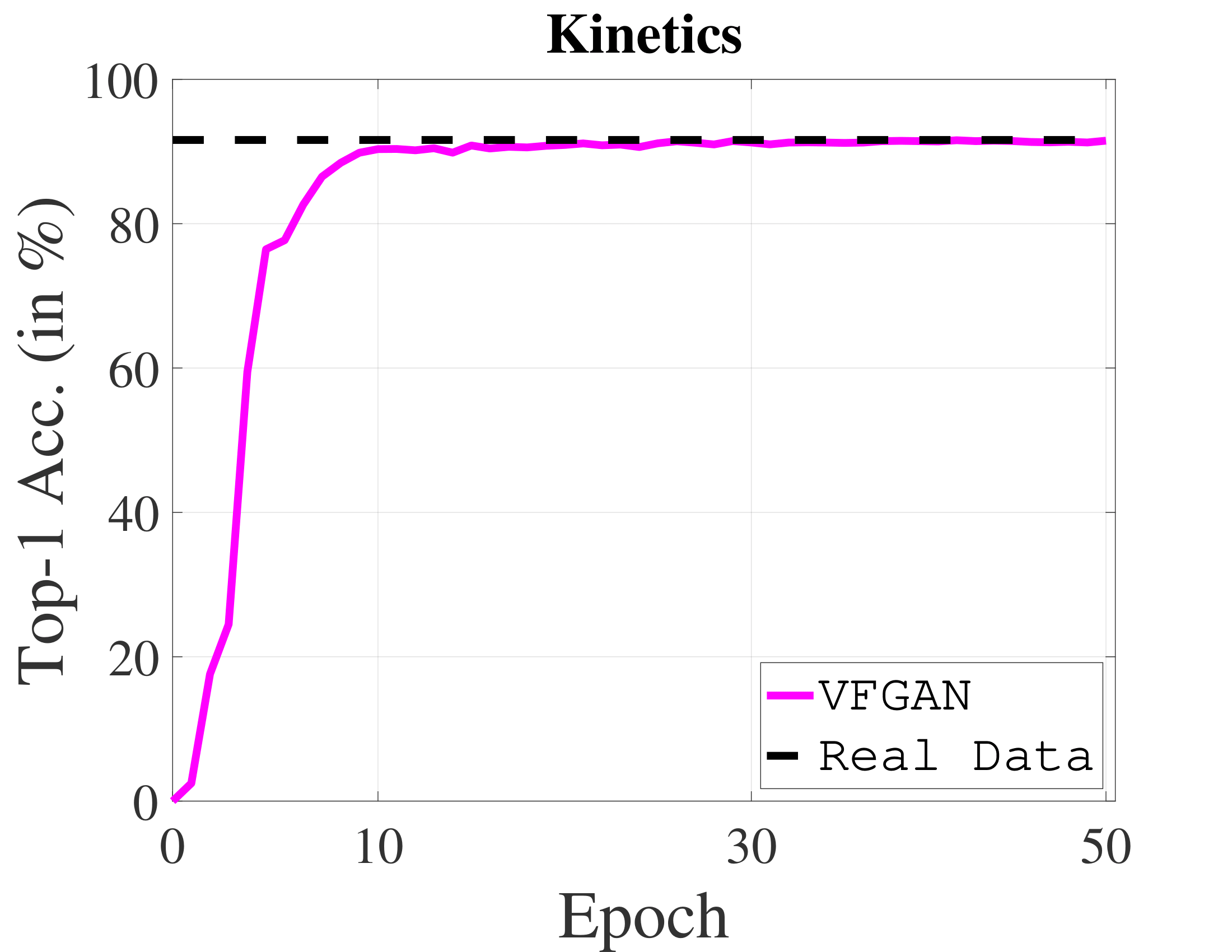}
	\includegraphics[width=.49\linewidth]{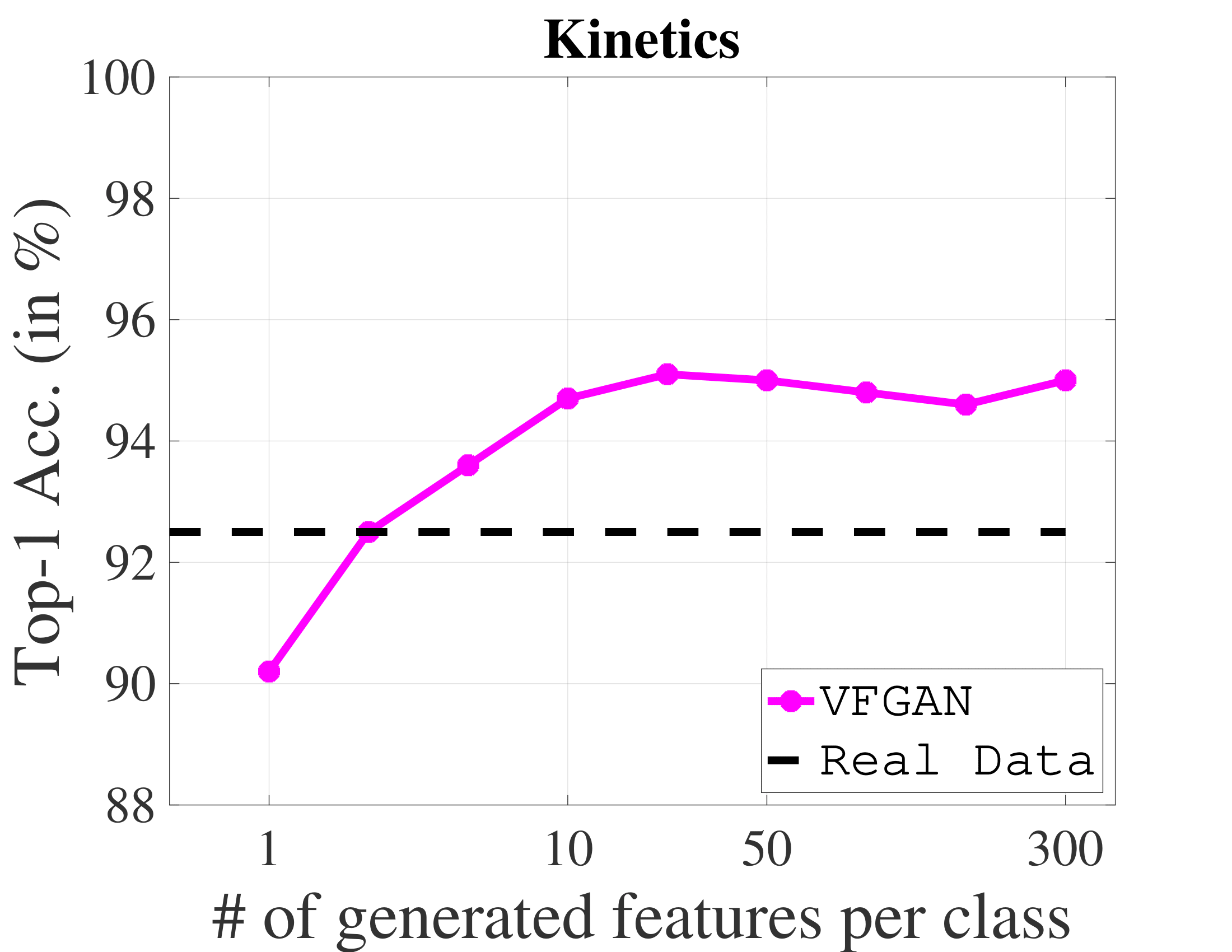} \\
	\includegraphics[width=.49\linewidth]{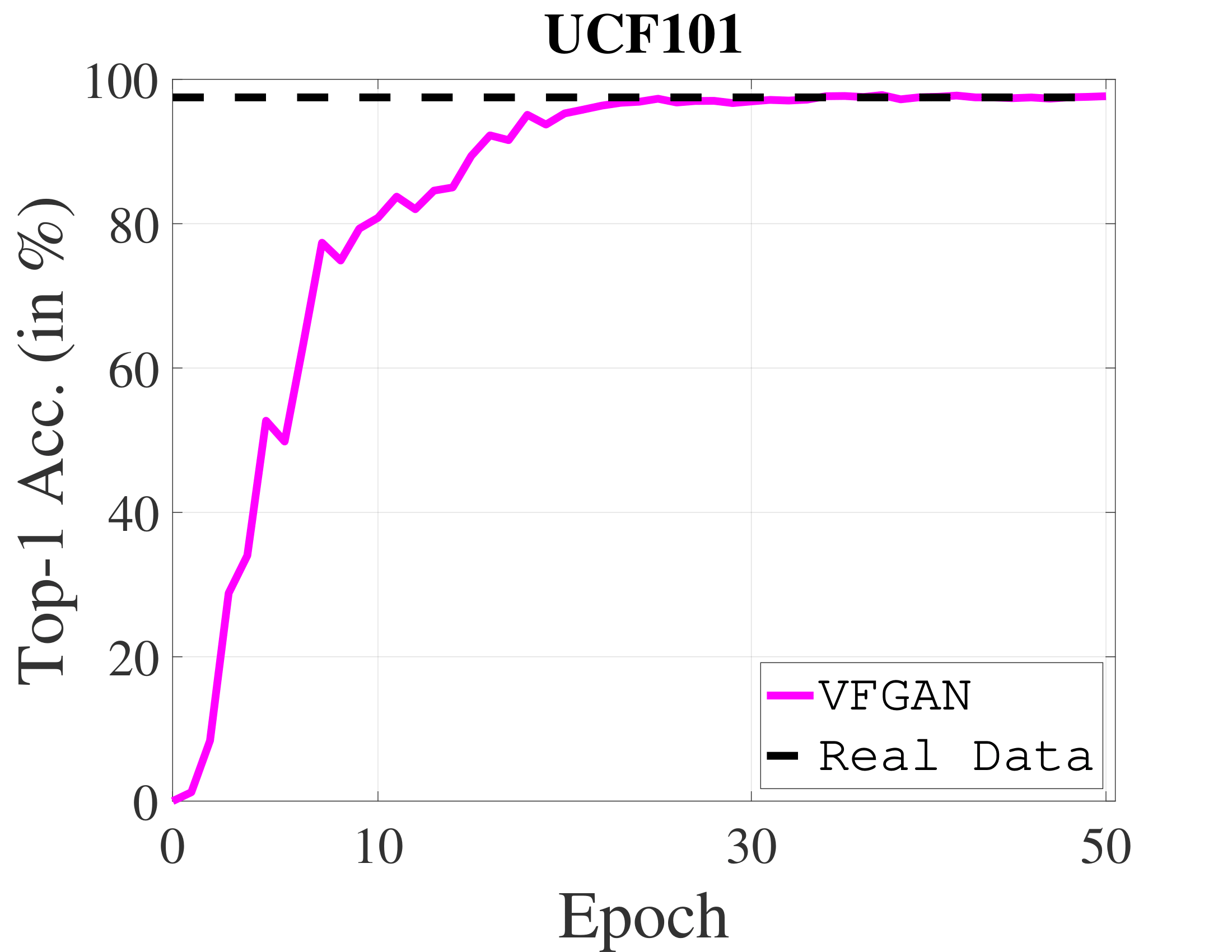}
	\includegraphics[width=.49\linewidth]{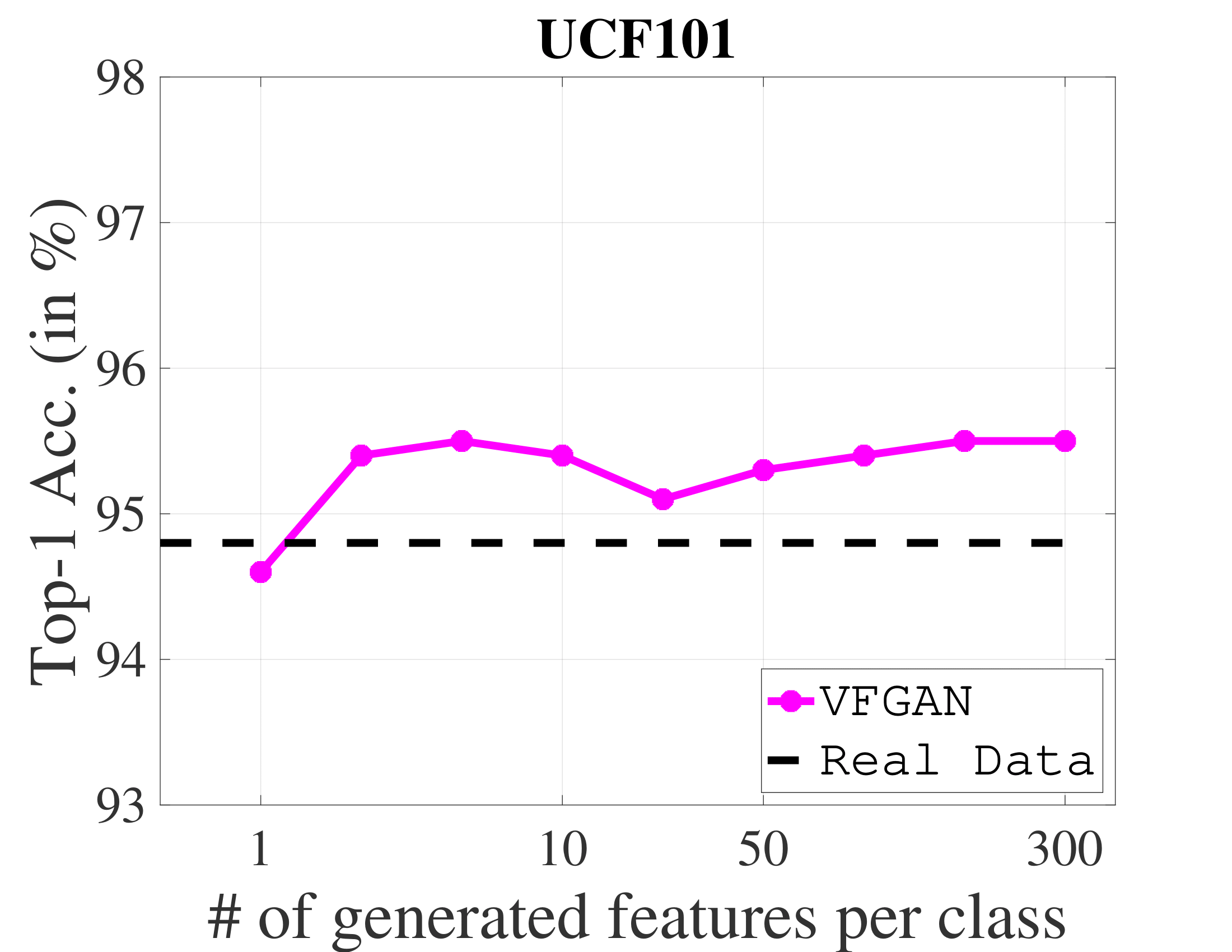} \\
	\includegraphics[width=.49\linewidth, trim=0 0 0 0,clip]{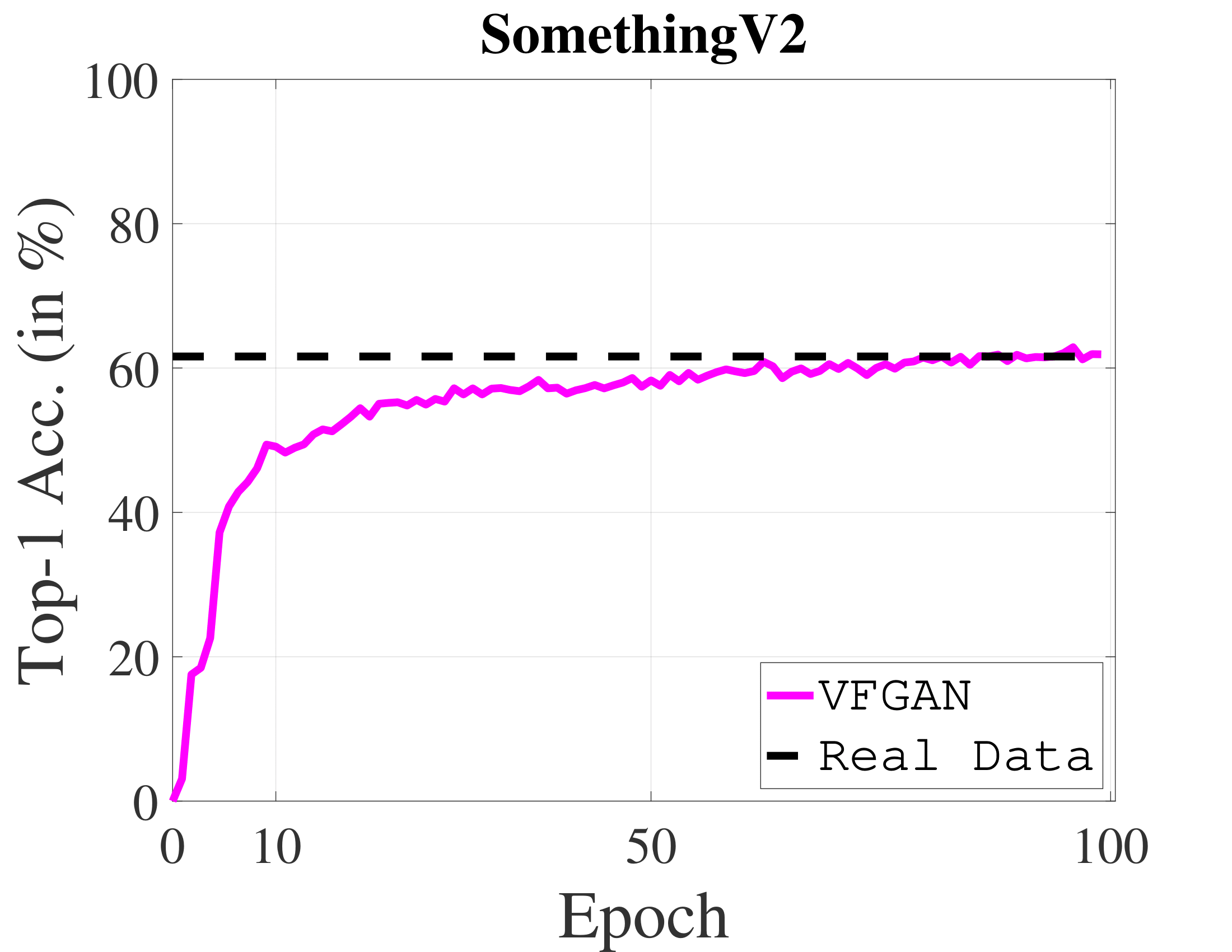}
	\includegraphics[width=.49\linewidth, trim=0 0 0 0,clip]{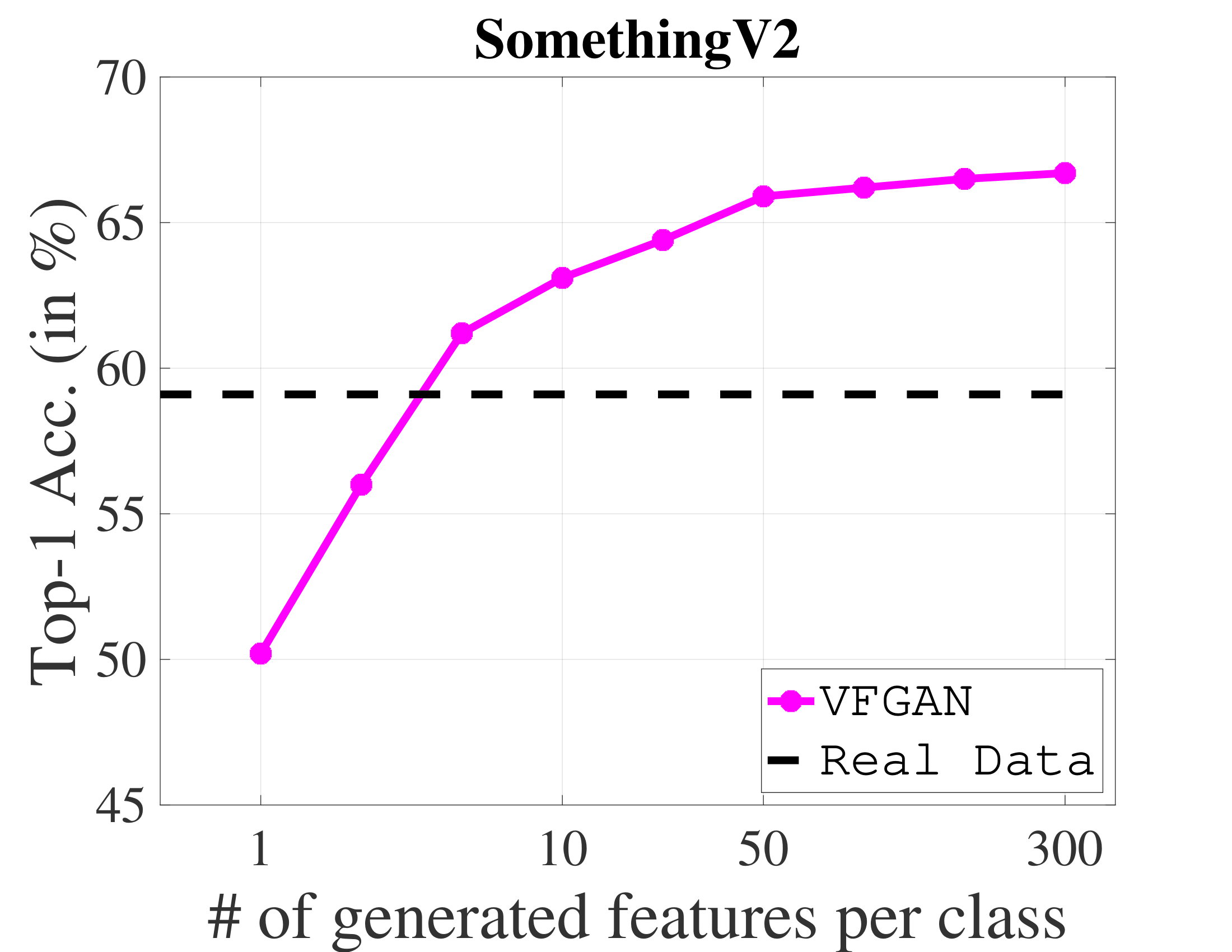}
	\caption{Model analysis of our VFGAN on Kinetics, UCF101 and SomethingV2. Left column: measuring the top-1 base class accuracy of the classifier trained with generated features w.r.t the VFGAN training epoch. Right column: increasing the number of generated VFGAN features per class w.r.t. novel class accuracy in the 1-shot 5-way setting.}
	\label{fig:gan_analysis}
\end{figure}

\subsection{Model Analysis: Video Feature Generation}
In this section, we evaluate the training stability of our VFGAN approach and the effect of the number of generated video features.

\myparagraph{Stability of VFGAN training.} GANs~\cite{GPMXWDOCB14} are known to be hard to train. In this experiment, we study the stability of our VFGAN by evaluating how well the generative model fits the real features of base classes, which occupies most of the training data used for learning the VFGAN. Instead of using Parzen window-based loglikelihood~\cite{GPMXWDOCB14} that is unstable, we train a softmax classifier with generated video features of base classes and report the classification accuracy on a held-out test set of the same classes. The left column in Figure~\ref{fig:gan_analysis} shows the classification accuracy w.r.t the number of GAN training epochs. On all three datasets, we observe a stable training trend and the generated features from our VFGAN almost reach the real feature upper bound. This indicates that our VFGAN fits the training data well and there is no stability issue of training VFGAN. We argue that the nice property of stable training can be attributed to the following two design choices: 1) we chose to generate video features instead of raw videos, which is a much harder task 2) we adopt the WGAN~\cite{arjovsky2017wasserstein} that is more stable than the vanilla GANs~\cite{GPMXWDOCB14}.

\myparagraph{Influence of the number of generated video features.} After we confirm the stability of our VFGAN on base classes, we study the generalization ability of VFGAN to generate video features of novel classes. 
We first train our VFGAN on the abundant training features of base class $X_b$ and a few training features of novel classes $X_n$. We then generate video features for the novel classes and train a softmax classifier using those generated video features. We vary the number of generated features to study how this relates to the novel class accuracy. The right column in Figure~\ref{fig:gan_analysis} shows that increasing the  
the number of generated features from 1
to 300 leads to a significant boost of accuracy, e.g. $90.2\%$ to $95.0\%$ on Kinetics and $50.2\%$ to $66.7\%$ on SomethingV2. In addition, across all three datasets, we observe that the classifiers trained with our generated features outperform the baseline trained with only 1-shot real training video~(the black dash line in right column of Figure~\ref{fig:gan_analysis}). The performance gain on UCF101 is not significant because the accuracy is almost saturated e.g., the baseline with only 1-shot training video already achieves $94.8\%$. These results indicate that our VFGAN is able to generate video features that are complementary to the few-shot real features of novel classes.

\section{Conclusion}

%We proposed a simple two-stage fine-tuning approach for few-shot object detection. Our method outperformed the previous meta-learning methods by a large margin on the current benchmarks. In addition, we built more reliable benchmarks with revised evaluation protocols. On the new benchmarks, our models achieved new states of the arts, and on the LVIS dataset our models improved the AP of rare classes by 4 points with negligible reduction of the AP of frequent classes.

In this work, we point out the importance of video feature learning for the few-shot video classification problem. We show that a simple two-stage baseline based on a 3D CNN outperforms prior best methods by a wide margin. We further improve the baseline by proposing two novel approaches: one  leverages the tag-labeled videos from YFCC100M using text-supported and video-based retrieval. The other one learns a generative adversarial networks that generate video features of novel classes from their semantic embeddings. Our approaches have obtained superior performance that almost saturates the existing Kinetics benchmark. Hence, we propose more challenging experimental settings, namely generalized few-shot video classification (GFSV) and many-way few-shot video classification that involves more novel classes. The extensive results on the new challenging benchmark demonstrate the effectiveness of our retrieval and feature generation approaches. 

\section*{Acknowledgments}
This work has been partially funded by the ERC 853489 - DEXIM and by the DFG – EXC number 2064/1 – Project number 390727645.

{
\small
	\bibliographystyle{IEEEtran}
	\bibliography{egbib}
}

\begin{IEEEbiography}[{\includegraphics[width=1in,height=1.2in,clip,keepaspectratio]{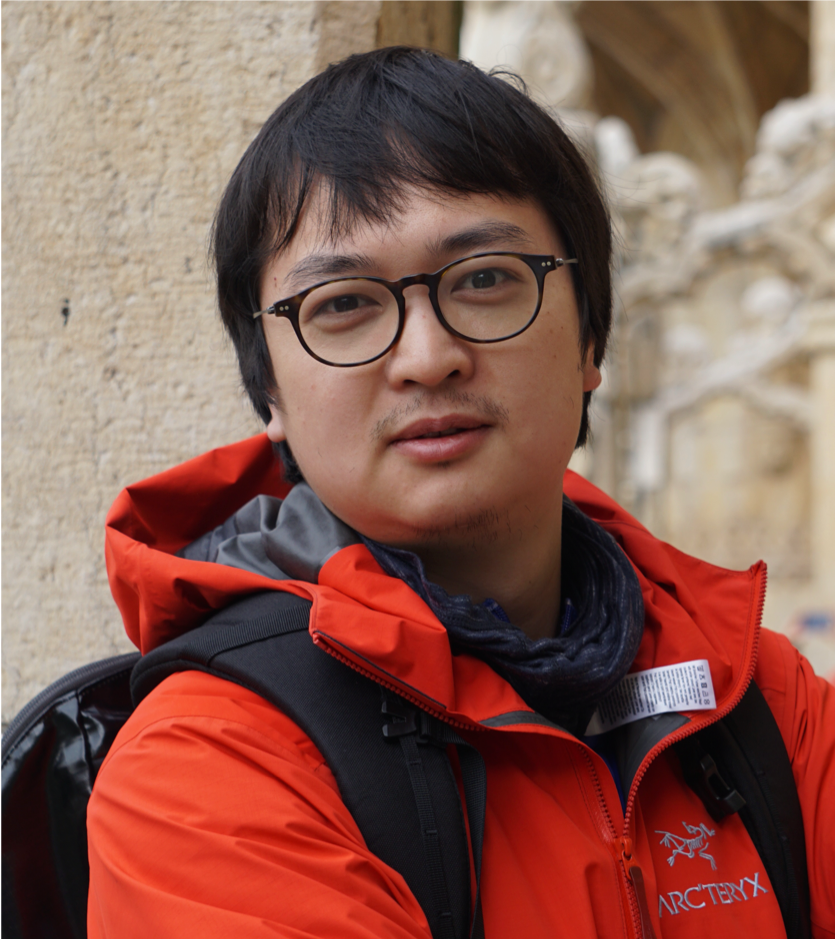}}]{Yongqin Xian} is a post-doctoral researcher at ETH Zurich, Switzerland. He received a bachelor degree from Beijing Institute of Technology (China) in 2013, a M.Sc. degree with honors from Saarland University (Germany) in 2016 and PhD degree (summa cum laude) from the Max Planck Institute for Informatics (Germany) in 2020. His research interests include zero-shot and few-shot learning for computer vision tasks.
\end{IEEEbiography} 

\vspace{-10mm}

\begin{IEEEbiography}[{\includegraphics[width=1in,height=1.2in,clip,keepaspectratio]{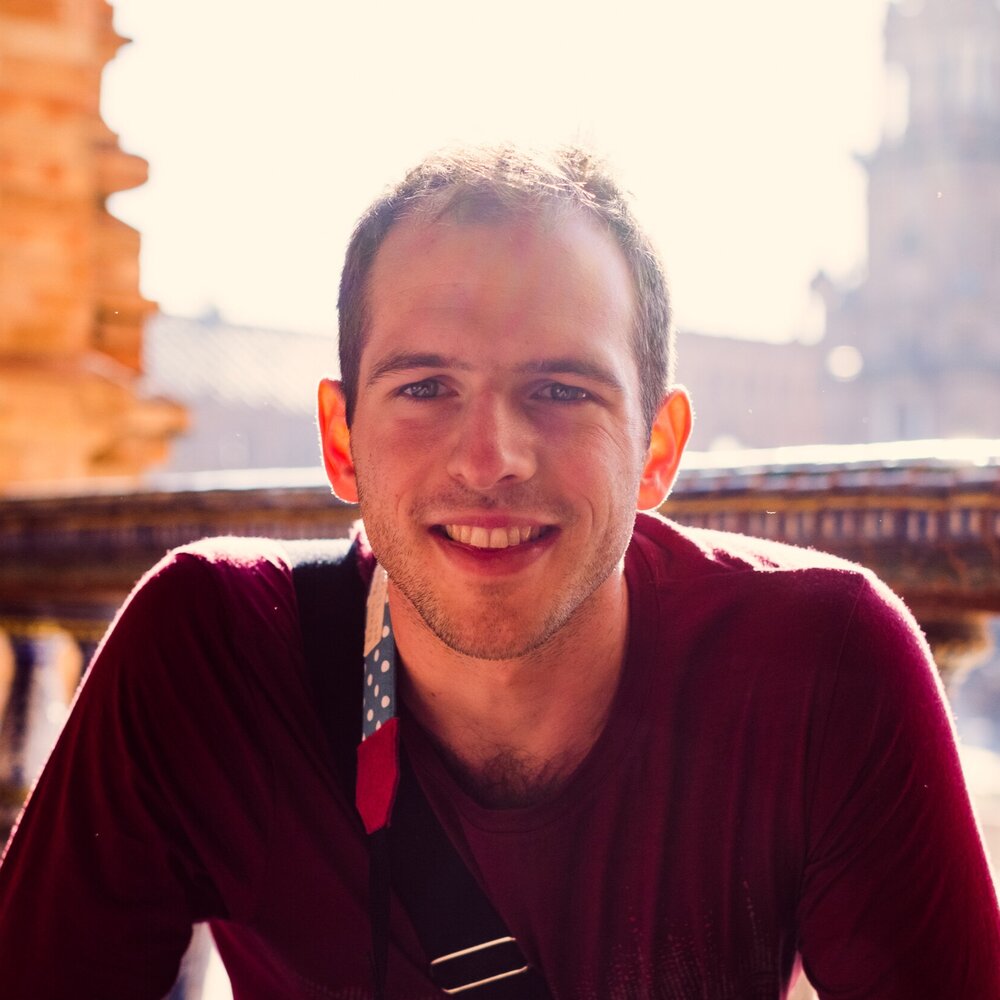}}]{Bruno Korbar} is a DPhil candidate at the University of Oxford (UK). He received a high-honours degree in computer science from Dartmouth College (2018). Prior to his current position, he was a research engineer at Facebook AI. His research interests include multi-modal video understanding and self-supervised learning. 

\end{IEEEbiography} 

\vspace{-10mm}

\begin{IEEEbiography}[{\includegraphics[width=1in,height=1.2in,clip,keepaspectratio]{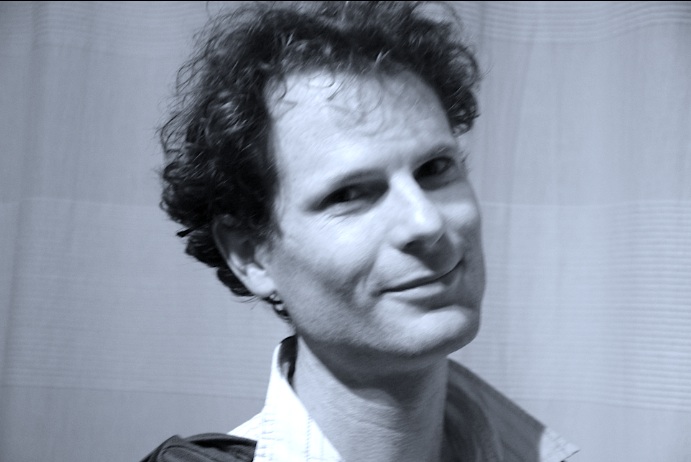}}]{Matthijs Douze} is a scientist at Facebook AI Research in Paris. He is working on large-scale indexing, machine learning with graphs and similarity search on images and videos. He holds a MSc and a PhD from ENSEEIHT (2004). He worked at INRIA Grenoble (2005-2015) on image indexing, large-scale vector indexing, event recognition in videos. He managed Kinovis, a large 3D motion capture studio  and developed high-performance geometric algorithms for constructive solid geometry operations. 
\end{IEEEbiography} 

\vspace{-10mm}

\begin{IEEEbiography}[{\includegraphics[width=1in,height=1.2in,clip,keepaspectratio]{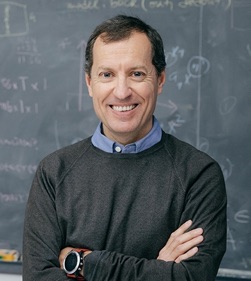}}]{Lorenzo Torresani} is a Professor in the Computer Science Department at Dartmouth College and a Research Scientist at Facebook AI Research (FAIR). He received a Laurea Degree in Computer Science with summa cum laude honors from the University of Milan (Italy) in 1996, and an M.S. and a Ph.D. in Computer Science from Stanford University in 2001 and 2005, respectively. In the past, he has worked at Microsoft Research, Like.com and Digital Persona. His research interests are in computer vision and deep learning. He is the recipient of several awards, including a CVPR best student paper prize, a National Science Foundation CAREER Award, a Google Faculty Research Award, three Facebook Faculty Awards, and a Fulbright U.S. Scholar Award.
\end{IEEEbiography} 

\vspace{-10mm}

\begin{IEEEbiography}[{\includegraphics[width=1in,height=1.2in,clip,keepaspectratio]{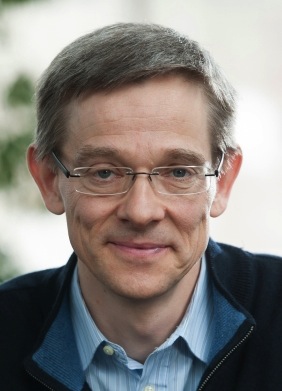}}]{Bernt Schiele} received masters degree in computer science from the University of Karlsruhe and INP Grenoble in 1994 and the PhD degree from INP Grenoble in computer vision in 1997. He was a postdoctoral associate and visiting assistant professor with MIT between 1997 and 2000. From 1999 until 2004, he was an assistant professor with ETH Zurich and, from 2004 to 2010, he was a full professor of computer science with TU Darmstadt. In 2010, he was appointed a scientific member of the Max Planck Society and director at the Max Planck Institute for Informatics. Since 2010, he has also been a professor at Saarland University. His main interests are computer vision, perceptual computing, statistical learning methods, wearable computers, and integration of multimodal sensor data. He is particularly interested in developing methods which work under real world conditions.
\end{IEEEbiography} 

\vspace{-10mm}

\begin{IEEEbiography}[{\includegraphics[width=1in,height=1.2in,clip,keepaspectratio]{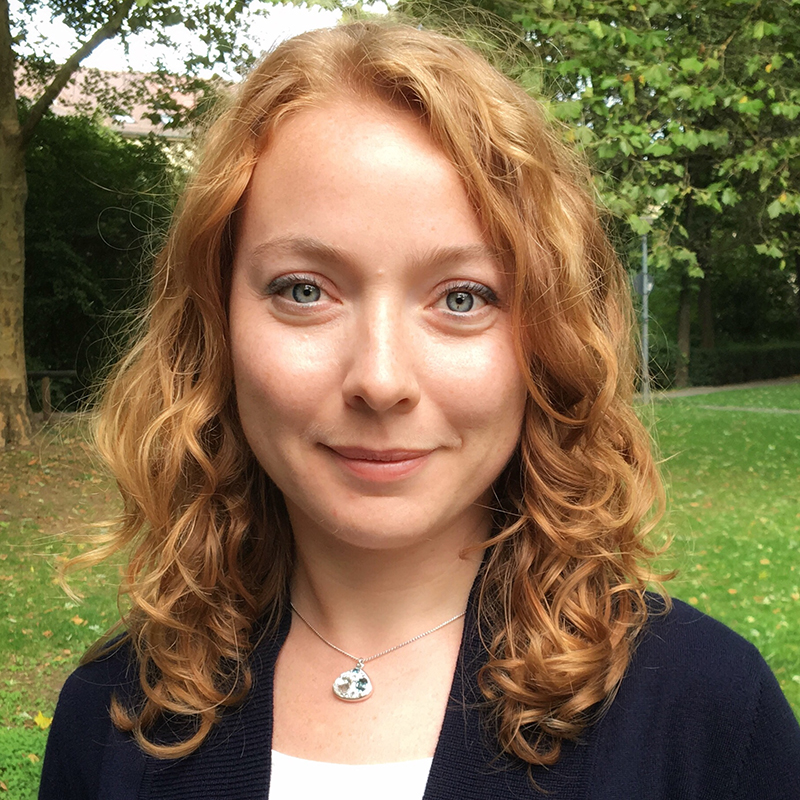}}]{Zeynep Akata} is a professor of Computer Science at the University of Tübingen. After her PhD in INRIA Rhone Alpes (France) in 2014, she worked as a post-doctoral researcher at the Max Planck Institute for Informatics (Germany) in 2014-2017, at UC Berkeley (USA) in 2016-2017 and as an assistant professor at the University of Amsterdam (The Netherlands) in 2017-2019. She received a Lise-Meitner Award for Excellent Women in Computer Science in 2014, an ERC Starting Grant in 2019 and the German Pattern Recognition Award in 2021. Her research interests include multimodal learning in low-data regimes such as zero- and few-shot learning as well as explainable machine learning focusing on vision and language.
\end{IEEEbiography}

\end{document}